\documentclass[10pt,twocolumn,letterpaper]{article}

\usepackage[pagenumbers]{cvpr}

\newcommand{\methodnamenoblank}{Murre}
\newcommand{\methodname}{\methodnamenoblank~}

\usepackage{booktabs} 
\usepackage{graphicx}
\usepackage{amsmath}
\usepackage{booktabs}
\usepackage{multirow}
\usepackage{colortbl}
\usepackage{makecell}
\usepackage{diagbox}
\usepackage{forloop}
\usepackage{bbding}
\usepackage{tabularx}
\usepackage{xcolor}
\usepackage{soul}
\usepackage{afterpage}
\usepackage{lipsum}
\usepackage{placeins}
\usepackage{float}

\definecolor{yellow}{rgb}{1,1, 0.6}
\definecolor{orange}{rgb}{1, 0.8, 0.6}
\definecolor{red}{rgb}{1, 0.6, 0.6}

\newcommand{\hlc}[2][yellow]{{%
    \colorlet{foo}{#1}%
    \sethlcolor{foo}\hl{#2}}%
}

\newcommand\blfootnote[1]{%
  \begingroup
  \renewcommand\thefootnote{}\footnote{#1}%
  \addtocounter{footnote}{-1}%
  \endgroup
}

\newcommand{\urlNewWindow}[1]{\href[pdfnewwindow=true]{#1}{\nolinkurl{#1}}}

\definecolor{myred}{rgb}{0.8,0,0}
\definecolor{mygreen}{rgb}{0,0.8,0}
\usepackage{pifont}

\definecolor{colorfirst}{rgb}{.866,.945, 0.831} 
\definecolor{colorsecond}{rgb}{1, 0.98, 0.83} 
\definecolor{colorthird}{rgb}{0.76, 0.87, 0.92} 

\definecolor{cvprblue}{rgb}{0.21,0.49,0.74}
\usepackage[pagebackref,breaklinks,colorlinks,allcolors=cvprblue]{hyperref}

\def\paperID{*}
\def\confName{CVPR}
\def\confYear{2025}

\title{Multi-view Reconstruction via SfM-guided Monocular Depth Estimation}

\author{
    Haoyu Guo$^{1*}$
    \quad He Zhu$^{2*}$
    \quad Sida Peng$^{1}$
    \quad Haotong Lin$^{1}$
    \quad Yunzhi Yan$^{1}$ \\
    \quad Tao Xie$^{1}$
    \quad Wenguan Wang$^{1}$
    \quad Xiaowei Zhou$^{1}$
    \quad Hujun Bao$^{1\dagger}$ \\
    $^1$Zhejiang University \quad
    $^2$Beijing Normal Univeristy \quad
}

\begin{document}

\twocolumn[
    \maketitle
    \begin{center}
    \captionsetup{type=figure}
    \includegraphics[width=\textwidth]{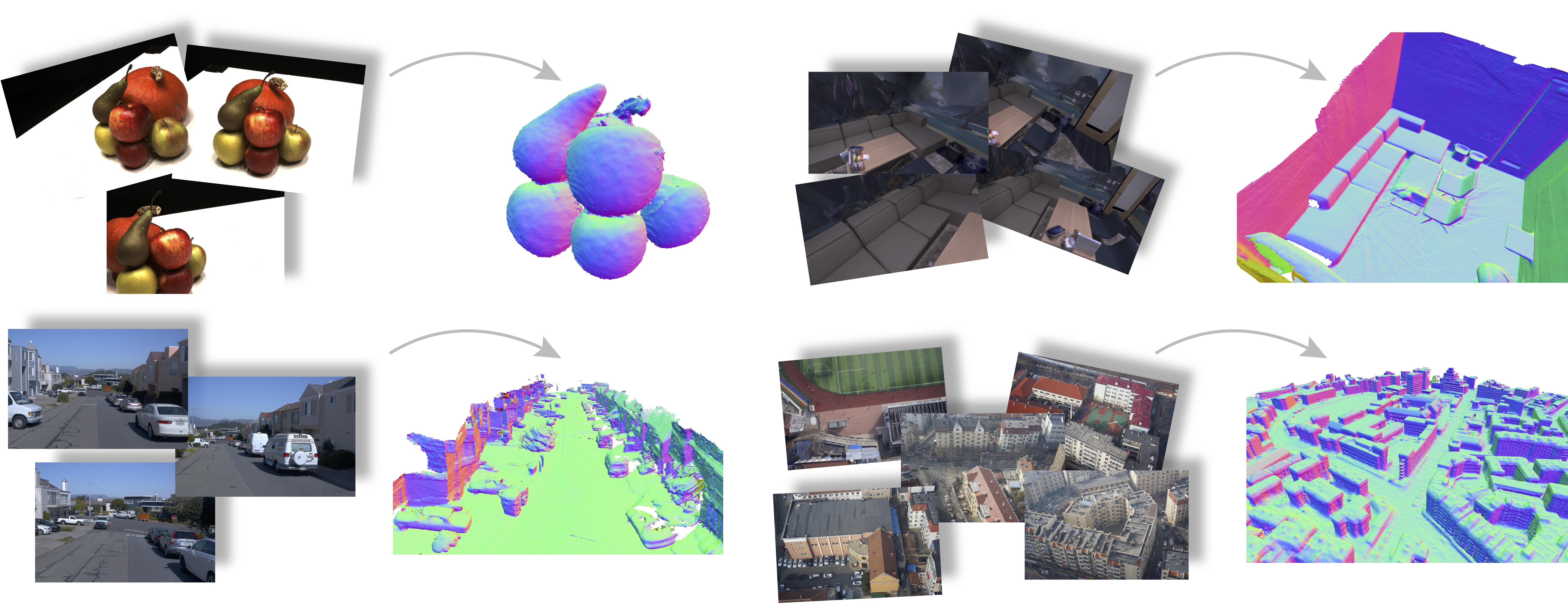}
    \captionof{figure}{%
    We propose \methodnamenoblank, a new method for multi-view 3D reconstruction based on SfM-guided monocular depth estimation. Building on Stable Diffusion~\cite{rombach2021highresolution}, \methodname demonstrates exceptional generalization capabilities after fine-tuning on a modest amount of synthetic data. \methodname is capable of achieving high-quality reconstructions for a variety of real-world scenarios, including object-level, indoor, street, and aerial scenes.
    }
    \label{fig:teaser}
\end{center}

    \bigbreak
]

\blfootnote{$^{*}$ Equal contribution}
\blfootnote{$^{\dagger}$ Corresponding author}
\begin{abstract}

This paper aims to reconstruct the scene geometry from multi-view images with strong robustness and high quality.
Previous learning-based methods incorporate neural networks into the multi-view stereo matching and have shown impressive reconstruction results.
However, due to the reliance on matching across input images, they typically suffer from high GPU memory consumption and tend to fail in sparse view scenarios.
To overcome this problem, we develop a new pipeline, named \methodnamenoblank, for multi-view geometry reconstruction of 3D scenes based on SfM-guided monocular depth estimation.
For input images, \methodname first recovers the SfM point cloud that captures the global scene structure, and then use it to guide a conditional diffusion model to produce multi-view metric depth maps for the final TSDF fusion.
By predicting the depth map from a single image, \methodname bypasses the multi-view matching step and naturally resolves the issues of previous MVS-based methods.
In addition, the diffusion-based model can easily leverage the powerful priors of 2D foundation models, achieving good generalization ability across diverse real-world scenes.
To obtain multi-view consistent depth maps, our key design is providing effective guidance on the diffusion model through the SfM point cloud, which is a condensed form of multi-view information, highlighting the scene's salient structure, and can be readily transformed into sparse depth maps to drive the image-space estimation process.
We evaluate the reconstruction quality of \methodname in various types of real-world datasets including indoor, streetscapes, and aerial scenes, surpassing state-of-the-art MVS-based and implicit neural reconstruction-based methods.
The code and supplementary materials are available at \urlNewWindow{https://zju3dv.github.io/murre/}.
\end{abstract}
\section{Introduction}

Multi-view 3D reconstruction aims to recover the accurate scene geometry from multiple images with known camera poses, which has extensive applications in robotics, autonomous driving, and virtual reality.
Traditional multi-view stereo algorithms \cite{schoenberger2016sfm,schoenberger2016mvs,stereopsis2010accurate,schonberger2016pixelwise} and neural reconstruction methods \cite{wang2021neus,yariv2021volume,yu2022monosdf} define an optimization target based on multi-view photometric consistency to reconstruct 3D scenes from the image signal.
Though these methods achieve impressive performance, they struggle with reconstruction in low-textured regions, as there might be several plausible solutions satisfying the photometric constraint in such area.
Moreover, the reconstruction speed of these methods is typically slow due to the time-consuming optimization process.

With the development of deep learning techniques, some recent methods \cite{sun2021neuralrecon} introduced neural networks into the multi-view stereo reconstruction framework for incorporating the 3D scene prior that is learned from the large-scale data.
These methods generally extract informative features from images, aggregate multi-view features in 3D space, and determine whether a point is on the 3D surface by implicitly considering the multi-view feature consistency.
With the learned scene prior, these methods resolve the ambiguity in low-textured regions and enable the feedforward inference with a faster reconstruction speed.

However, such learning-based MVS methods \cite{yao2018mvsnet,yao2019recurrent,gu2020cascade,xu2023iterative,wang2022mvster,yang2022mvs2d,cao2022mvsformer} still face three challenges.
First, due to the high spatial complexity of 3D space, aggregating features in 3D space tends to consume a large amount of GPU memory, thereby limiting the resolution of the reconstructed geometry.
Second, these methods implicitly rely on the inductive bias of multi-view consistency, thus tending to fail in a sparse view setting where there are many regions that cannot be matched across input images.
Third, these methods generally require ground-truth 3D data for training. Because high-quality 3D data is limited, it is difficult to train them on a large amount of data, resulting in a limited generalization ability.

In this paper, we propose a novel learning-based framework, named \methodnamenoblank, that combines SfM techniques with monocular depth estimators to reconstruct 3D geometry from multi-view images.
Our core innovation lies in injecting SfM priors into diffusion-based depth estimation to produce multi-view consistent depth maps for input images, achieving 3D reconstruction with strong generalization capability and high quality.
Specifically, given input images, \methodname first recovers a sparse point cloud through an off-the-shelf SfM method \cite{schonberger2016structure,he2024dfsfm}, which captures the scene structure globally.
Then, the point cloud acts as a guidance for a conditional diffusion model, which takes the projected sparse depth map and an RGB image as input and predicts the corresponding metric depth map.
Finally, the 3D geometry of the target scene is obtained by performing the point cloud fusion \cite{merrell2007real,galliani2015massively} or TSDF fusion \cite{newcombe2011kinectfusion} on metric depth maps of input images.
\methodname has two advantages over previous learning-based MVS methods \cite{yao2018mvsnet,yao2019recurrent,gu2020cascade,xu2023iterative}.
First, by utilizing an SfM-guided diffusion model to predict the depth map from a single image, \methodname bypasses the multi-view matching step, thereby addressing the issues of high memory consumption and difficulty in handling sparse view scenarios.
Second, \methodname adapts the diffusion model from a pre-trained stable diffusion model, making a clever use of the powerful priors of the 2D foundation model.
Our experimental results in \Cref{sec:exp} show that fine-tuning on a small amount of synthetic 3D data can achieve good generalization results in diverse real-world scenes, including object-level, indoor, street and aerial scenes.

More specifically, a key design of \methodname is exploring the explicit intermediate representation, SfM point clouds, to guide the conditional diffusion model to produce multi-view consistent depth maps with an accurate scale.
As a condensed form of multi-view information, SfM point clouds effectively highlight the scene's salient structure and can be readily transformed into sparse depth maps, which serve as conditional signal well-suited for the image-space estimation process.
When predicting the depth map of an input image, we first project the reconstruction SfM point cloud onto the corresponding view.
To perform the dense guidance on the diffusion model, we densify the sparse depth map using the $k$-nearest neighbors algorithm and compute a corresponding distance map as a confidence indicator.
The diffusion model takes the RGB image, densified depth map, and distance map as input, and predicts the target metric depth map.
Experimental results show that our strategy significantly enhances the multi-view consistency of the predicted depth maps from the diffusion model.

We evaluate \methodname on several reconstruction benchmarks of diverse types, including DTU~\cite{jensen2014large}, ScanNet~\cite{dai2017scannet}, Replica~\cite{replica19arxiv}, Waymo~\cite{Sun_2020_CVPR} and UrbanScene3D~\cite{UrbanScene3D} datasets. 
\methodname consistently outperforms state-of-the-art monocular depth estimation~\cite{ke2023repurposing,yang2024depth,yang2024depthv2} methods and achieves competitive performance with recent Multi-view Stereo and implicit neural reconstruction-based method~\cite{yu2022monosdf,sun2021neuralrecon,sayed2022simplerecon}. 
Detailed ablations and analysis on the SfM method and depth conditioning designs are included to demonstrate the robustness and efficiency of our design.

In summary, our contributions are as follows:
\begin{itemize}
    \item We propose a novel approach to inject SfM priors into diffusion-based depth estimation, enabling highly accurate and multi-view consistent depth predictions for each viewpoint.
    \item Based on the proposed depth estimator, we design a new multi-view 3D geometry reconstruction framework and process some synthetic datasets to facilitate training.
    \item We evaluate our method on diverse real-world scene data, including objects, indoor, streetscapes, and aerial scenes, demonstrating the superior performance and generalization capability of our approach.
\end{itemize}
\section{Related works}

\subsection{Multi-view reconstruction}

Multi-view 3D reconstruction has been a long focus in the field of 3D computer vision for decades.
Traditional multi-view stereo (MVS) methods \cite{schoenberger2016sfm,schoenberger2016mvs,stereopsis2010accurate,schonberger2016pixelwise,fuhrmann2014mve,jancosek2011multi} primarily rely on photometric consistency to perform dense matching between images from different views to estimate depth, leading to a subsequent reconstruction of the scene. However, limited by their weak feature representation capability, they struggle to deliver high-quality reconstruction results in challenging scenes.

Recently, with the rise of deep learning, learning-based MVS methods have been widely explored. These methods typically construct a 3D cost volume and utilize 3D CNNs \cite{yao2018mvsnet,yao2019recurrent,gu2020cascade,xu2023iterative} or transformers \cite{wang2022mvster,yang2022mvs2d,cao2022mvsformer} to predict per-view depth map, and then fuse all views together. 
However, the introduction of 3D cost volume results in massive computational overhead and still struggles to provide reasonable reconstruction results under sparse viewpoints or textureless situations. 
In contrast, NeuralRecon \cite{sun2021neuralrecon} directly constructs a feature volume in the world coordinate system and predicts a TSDF volume for scene reconstruction. 
Although these approaches can handle low-textured regions, they require expensive 3D ground truth as supervision and lacks reconstruction details.
Optimization-based neural reconstruction approaches \cite{wang2021neus,yariv2021volume,yu2022monosdf,li2023neuralangelo} leverage RGB supervision to implicitly optimize the 3D scene representation, which alleviates the requirement for accurate 3D ground truth. However, they still necessitate time-consuming optimization for each distinct scene.
Our method proposes a new multi-view reconstruction framework which combines monocular depth estimation with sparse SfM point clouds, addressing the challenges posed by learning-based methods in handling sparse viewpoints and weak textures. 

\begin{figure*}[t]
\centering
\includegraphics[width=\linewidth]{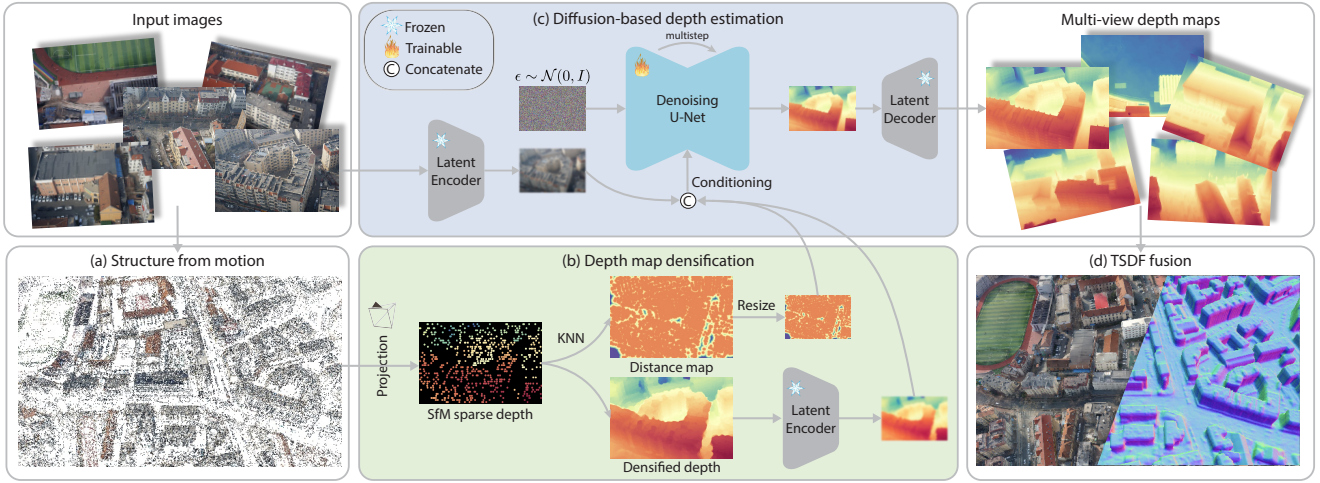}
\caption{\textbf{Overview of our multi-view reconstruction pipeline.} Given multi-view images, we first employ a Structure from Motion (SfM) method \cite{he2024dfsfm} to derive sparse 3D scene structures (a). These 3D structures are then encoded into an intermediate explicit representation (b), which is used as a condition for depth estimation (c). Finally, we conduct a TSDF fusion \cite{newcombe2011kinectfusion} to achieve the final reconstruction (d).}
\label{fig:pipeline}
\vspace{-10pt}
\end{figure*}

\subsection{Monocular depth estimation}

Monocular depth estimation, which aims to directly predict a dense depth map from one single image, holds expansive application value in areas such as generative AI \cite{rombach2021highresolution,hollein2023text2room} and autonomous driving \cite{liu2022vision,fonder2021m4depth}, specifically, they are shown to provide a strong prior for multi-view reconstruction \cite{yu2022monosdf,wei2021nerfingmvs,chung2024depth,turkulainen2024dn}.
However, recovering metric scale depth purely from one single image is an ill-posed problem. 
Previous metric monocular depth estimation methods \cite{bhat2021adabins,bhat2022localbins,yuan2203new,bhat2023zoedepth} learn the metric scale by training on specific scene data, but they essentially overfit on the training scenarios and face difficulties in generalizing to very different scenes.
Recent works \cite{facil2019cam,yin2023metric3d} boost the generalizability by explicitly modeling different camera models through encoding or constructing a canonical camera model. 
Although these methods improve the performance on depth estimation, they still struggle to produce multi-view consistent depth maps for observed images of a target 3D scene.

Alternatively, affine-invariant relative monocular depth estimation methods predict the relative relationships between pixels without a specific metric scale.
\cite{ranftl2020towards,yang2024depth,yang2024depthv2} achieve better generalization capabilities through pre-training on large-scale datasets.
Recently, diffusion-based methods \cite{ke2023repurposing,fu2024geowizard,gui2024depthfm} repurpose the diffusion \cite{ho2020denoising} framework to denoise the depth map from the image being used as the condition. With the help of the powerful priors provided by Stable Diffusion \cite{rombach2021highresolution}, they fine-tune models with decent generalization on a minor quantity of synthetic datasets. However, these methods lack global metric information and struggle to guarantee the internal relative scale consistency of a single depth map, making them less applicable for reconstruction.
Additionally, \cite{luo2020consistent,kopf2021robust,shao2024learning,yang2024depthvideo,hu2024depthcrafter} explore estimating consistent depth from monocular video.

\subsection{Depth completion}

Depth completion aims to recover dense pixel-wise depth maps from sparse depth maps obtained via SfM or sensors. Traditional methods have made extensive attempts to incorporate multi-modal data fusion into the completion pipeline, such as RGB images \cite{jeon2021abcd,metzger2023guided,qiu2019deeplidar,zhao2021adaptive}, semantic maps \cite{markert2022segmentation}, and normal maps \cite{xu2019depth}. Building upon it, later methods focus on refining the network architectures, such as different convolution structures \cite{cheng2018depth,cheng2020cspn++,lin2022dynamic,liu2022graphcspn}, and training strategies \cite{zhang2023completionformer}.
However, they neglect the inner properties of the non-uniform sparse depth points. \cite{conti2023sparsity,wu2022sparse,long2024sparsedc} try to address this by iteratively merging or locality-aware spatial propagation. Although they achieve quality improvement, these methods are all premised on the assumption of accurate sparse depth, and struggle to yield satisfactory results when tasked with noise-containing depths such as SfM depth.

\section{Method}

Given multi-view images of the target scene along with sparse reconstruction results from Structure from Motion (SfM), our goal is to utilize the SfM priors to estimate the dense depth for each view, which can be fused to obtain the geometry reconstruction of the scene, as illustrated in \Cref{fig:pipeline}.
We build our depth estimator based on a conditional diffusion model (\Cref{sec:diffusion}).
To leverage multi-view information, we propose an explicit intermediate representation to incorporate SfM priors during the depth estimation process (\Cref{sec:sfm-prior}).
Based on our proposed depth estimation method, we design a pipeline to reconstruct the scene geometry from input multi-view images (\Cref{sec:mv-recon}).
Finally, we elaborate on our training details in \Cref{sec:training}.

\subsection{Diffusion-based depth estimation}
\label{sec:diffusion}

Recent multi-view stereo methods \cite{yao2018mvsnet,sayed2022simplerecon,zhang2023geomvsnet} typically rely on multi-view image matching to build cost volumes, which makes them difficult to handle textureless regions and sparse viewpoints.
To address this, we propose injecting multi-view information into monocular depth estimation \cite{ke2023repurposing} to avoid the need for multi-view image matching.
To achieve this goal, we condition our method on a diffusion-based monocular depth estimation model, which has shown strong performance benefiting from powerful modern image diffusion models.
In the following, we first introduce the prior knowledge of the image diffusion model and then explain how to use it to facilitate depth estimation.

Image generation field has recently made significant progress with large-scale data and diffusion models. 
A representative method is Denoising Diffusion Probabilistic Models (DDPM)~\cite{ho2020denoising}, which learns the reverse process of adding noise and generating a clean image from a random noise.
To achieve high-resolution image generation, \cite{rombach2021highresolution} suggests conducting the noise addition and denoising process in the latent space. 
A Variational Autoencoder (VAE)~\cite{esser2021taming} is first trained on a very large dataset of images, enabling bidirectional transformations between image space and latent space.
\cite{saharia2022photorealistic} introduces a conditioning mechanism to diffusion models, enabling control over the generated outcomes based on textual or visual information.

Inspired by \cite{ke2023repurposing}, we formulate depth estimation as a latent conditional diffusion process.
Specifically, in the forward process, the Gaussian noise is gradually added to the depth $d$ to obtain the noisy depth sample $d_t$, starting from $d_0 = d$ to levels $t \in \{1, \ldots, T\}$ as the following equation:
\begin{equation}
d_t = \sqrt{\bar{\alpha}_t} d_0 + \sqrt{1 - \bar{\alpha}_t} \epsilon,
\end{equation}
where $\epsilon \sim \mathcal{N}(0, I)$, $\bar{\alpha}_t = \prod_{s=1}^t (1 - \beta_s)$, and $\{\beta_1, \ldots, \beta_T\}$ is the variance schedule of a process with $T$ steps. 
In the reverse process, a learned denoiser is used to predict the noise $\epsilon$ from the monocular image condition, which is fed into the denoising equation \cite{ho2020denoising} to calculate the depth map $d_{t-1}$ from $d_t$.
Finally, we iteratively perform the denoising process and obtain the clean latent $d_0$, which is converted into the depth through the decoder of the VAE.

\subsection{Injecting SfM priors into depth estimator}
\label{sec:sfm-prior}
Previous methods \cite{ke2023repurposing} achieve high-quality monocular depth estimation by using the diffusion method with monocular images as a condition.
However, directly fusing their output depth produces poor reconstruction results because the model struggles to predict multi-view consistent depth maps with only monocular images due to the ill-posedness.
To achieve high quality multi-view 3D reconstruction, an intuitive solution is to incorporate multi-view information into the depth estimation model.
However, directly incorporating information from multi views under the diffusion framework is quite challenging, because the number of input views and the distribution of their camera poses are unknown, resulting in a large variance in input data.
Moreover, the image-space estimation process makes it difficult to fuse multi-view image features.

To overcome these problems, we propose to use the SfM point cloud as an explicit intermediate representation, enabling effective and efficient injection of multi-view information into the monocular depth estimation model.
Using the SfM point cloud as a condition has several advantages: Firstly, SfM points can be regarded as a condensed form of multi-view information and scene 3d structure, which can guide the prediction of the depth maps. Secondly, the research on SfM has been conducted for nearly two decades, making it highly mature and robust \cite{ng2003sift,schonberger2016structure,sun2021loftr,he2024dfsfm}.
Finally, SfM point clouds can be projected into each view according to the visibility as sparse depth maps, making them well-suited for the image-space estimation task.

Since the projected depth maps are sparse and not suitable for image-space conditioning, we propose to first densify them through the $k$-nearest neighbors (KNN) interpolation.
Specifically, for each pixel $p$ in a sparse depth map that lacks a value, we find the $k$ nearest pixels with values based on Euclidean distance. 
We then perform a weighted average of these $k$ pixel depths using the inverse of the distances as weights to assign a value to pixel $p$. 
Using this method, we densify the depth map. 
After that, we pass the densified depth map through an encoder to serve as a condition.
To help the network to distinguish between original and interpolated pixels and better utilize the SfM information, we use an additional map that records the Euclidean distance from each pixel to its nearest neighbor with a value. 
The distance map is used as an additional condition.

Due to the significant variations in the range of SfM-projected depth values across different views of various scenes, we first normalize the spase SfM depth maps.
Considering that SfM points often contain outliers, we initially filter out the top and bottom $2\%$ of depth values for each view.
Given that SfM projections only cover a small portion of the pixels and the actual depth range of the entire image maybe greater than the depth range of the SfM points, we expand the calculated range based on SfM depth. 
Specifically, we adjust the minimum to 0.8 times its original value and the maximum to 1.2 times. 
This adjustment ensures that the entire image's depth range is well covered in most cases, thereby minimizing the impact of edge cases on training.
For the ground truth depth map, we use the same range for normalization and then map it to the latent space to supervise the denoising process. 
This design ensures that the trained model can predict depths on the same scale as the input SfM depth, which can then be effectively applied to subsequent multi-view 3D reconstruction.

\subsection{Multi-view 3D reconstruction}
\label{sec:mv-recon}

Given input multi-view images, we first reconstruct sparse point clouds using SfM techniques.
Based on the SfM point cloud, we predict depth maps for the input images.
Specifically, we map the image and SfM information used as conditions to the latent space following the method described in \Cref{sec:sfm-prior} to obtain the normalized depth map.
We use the recorded normalization range of the SfM depth to de-normalize the predicted depth back to the original scale of the SfM depth.
After that, we use off-the-shelf multi-view depth fusion methods such as point cloud fusion \cite{merrell2007real,galliani2015massively} or TSDF fusion \cite{newcombe2011kinectfusion} to obtain the scene’s 3D geometry.

We found that directly fusing the predicted depth maps yields suboptimal results.
One reason is a slight inconsistency in the scale between the predicted depth and the SfM depth.
To overcome this issue, we further align the predicted depth map with the SfM sparse depth using RANSAC based linear regressor.
Specifically, RANSAC iteratively identifies inliers and estimates the optimal scale and shift parameters by minimizing the reprojection error between the predicted depth and the SfM depth. In each iteration, a subset of points is randomly sampled to compute a candidate transformation, which is then evaluated based on the number of inliers it produces. The final transformation is selected as the one with the highest consensus, ensuring robust alignment by effectively handling outliers and adjusting for both translation and scaling discrepancies.

\subsection{Training}
\label{sec:training}

To train our model, we first run the SfM algorithm to reconstruct sparse point clouds of training scenes from input multi-view images.
Specifically, we choose Detector-free SfM~\cite{he2024dfsfm}, which is based on a detector-free matcher~\cite{sun2021loftr,edstedt2023dkm}. 
This method is better suited for handling areas with weak textures compared to classical SfM approaches \cite{schonberger2016structure}.
After running SfM on each scene, we project the obtained sparse point cloud according to the visibility to get the sparse depth map for each view.

Based on the constructed training data, we train a depth estimator that uses both images and SfM depth as conditions.
To enable robust generalization capabilities, instead of training from scratch, we use the pretrained Stable Diffusion V2 model \cite{rombach2021highresolution} as the initialization, and we fix the VAE and only finetune the UNet during the training.
We normalize the GT depth map of each frame according to \Cref{sec:sfm-prior}, then replicate it across three channels to use the VAE's encoder to map it into the latent space.
We then add noise to the latent code of the GT depth and train the denoiser based on this.
In addition to the latent codes of the image and densified SfM depth map, we also utilize the Euclidean distance map as condition. Because the information of distance map is of low frequency, instead of passing it through the encoder, we directly resize it to the same resolution as the latent code.
We concatenate them together with the noise and send them into the UNet to predict $\hat{\epsilon}$. The training loss is an mse loss defined as: 
\begin{equation}
    \mathcal{L} = \left\| \epsilon - \hat{\epsilon} \right\|^2,  \label{eq:1}
\end{equation}
where $\epsilon$ is the noise added to the latent code of GT depth.

\section{Implementation details}
We employ the official pre-trained model of Stable Diffusion v2 model \cite{rombach2021highresolution} as our foundation, with the Variational Autoencoder (VAE) fixed and only the U-Net fine-tuned. Our framework is implemented using PyTorch Lightning \cite{Falcon2019} and diffusers \cite{von-platen-etal-2022-diffusers}. During the training stage, we utilize a mixture of data from Hypersim and 3D Ken Burns, and for alignment purposes, we downscale the resolution of Hypersim from $768\times1024$ to $384\times512$, and center-cropped the 3D Ken Burns data from $512\times512$ to $384\times512$. We train our model using eight Nvidia Tesla A100 GPUs with a batch size of 128 across 50 epochs (about 25 hours), employing the Adam optimizer with a learning rate of $3 \times 10^{-5}$.
To enhance the robustness of depth estimation, inspired by \cite{ke2023repurposing}, we employ a test-time ensemble approach, performing inference 5 times and then taking the pixel-wise median.

\section{Experiments}
\label{sec:exp}

\subsection{Datasets and metrics}

\subsubsection{Training datasets}
We trained our model on two synthetic datasets.
(1) \textbf{Hypersim} \cite{roberts:2021} is a photorealistic synthetic dataset consisting of 461 indoor scenes. Based on the official split, we filter out broken or low-quality scenes from the 365 training scenes, retaining 345 scenes with 56,077 frames. For each scene, we performed triangulation based on the provided camera parameters to obtain sparse reconstruction results.
(2) \textbf{3D Ken Burns} \cite{niklaus20193d} is a large-scale synthetic dataset, containing 23 in-the-wild scenes with two sequences each. Due to the unknown camera parameters in this dataset, we employ Detector-free SfM \cite{he2024dfsfm} for calibration and sparse reconstruction. After excluding sequences of poor quality or those where SfM does not work properly, we retained 12 sequences with 30,309 frames for training.

\subsubsection{Evaluation datasets.}

To validate the generalizability and robustness of our multi-view reconstruction method across different data types, we conduct evaluations on various types of real-world data.
(1) {Object-level scenes}: \textbf{DTU} \cite{jensen2014large} DTU comprises 80 object-level scenes captured using multiple cameras under various lighting conditions. We use 15 scenes for 3-view reconstruction evaluation, following \cite{yu2022monosdf}.
(2) {Indoor scenes}: \textbf{ScanNet} \cite{dai2017scannet} is a RGBD dataset containing 1,613 indoor scenes. We evaluate on 4 scenes in validation set, following \cite{yu2022monosdf,guo2022manhattan}.
\textbf{Replica} \cite{replica19arxiv} includes 18 indoor scenes with high quality Ground Truth geometry at room scale. We evaluate on 3 test scenes according to the split in \cite{yu2022monosdf}.
(3) {Streetscapes}: \textbf{Waymo Open Dataset} \cite{Sun_2020_CVPR} is an autonomous driving dataset collected with five LiDAR sensors and five high-resolution pinhole cameras. We select 32 static sequences for evaluation, following \cite{guo2023streetsurf}.
(4) {Aerial views}: \textbf{UrbanScene3D} \cite{UrbanScene3D} is an outdoor dataset captured using drones, each scene with 2500-6000 images. We evaluate on two scenes covering $10^6$ square meters.
For DTU, ScanNet, Replica, and Waymo, we conduct sparse reconstructions using the triangulation mode of Detector-free SfM from known camera parameters. For UrbanScene3D, we obtain the calibration as well as sparse reconstruction results by COLMAP.

\begin{table*}[t]
\centering
\setlength{\tabcolsep}{1.5pt}
\captionof{table}{\textbf{Quantitative comparison on DTU dataset}. We evaluate Chamfer Distance of each method on each scene (measured in millimeters) and highlight the top three methods in \hlc[red]{red}, \hlc[orange]{orange}, and \hlc[yellow]{yellow}, respectively. Apart from metrics, we also report the amount of training data used by each method. Please refer to the supplementary materials for results on other datasets.}
\vspace{-5pt}
\begin{tabular}{ccccccccccccc}
\toprule
& \footnotesize{COLMAP} & \scriptsize{RealityCapture} & \small{VolSDF} & \small{MonoSDF} & \small{Marigold} & \tiny{Depth-Anything} & \tiny{Depth-Anythingv2} & \small{Metric3D} & \small{MVSNet} & \scriptsize{IGEV-MVS} & \small{DUSt3R} & \small{Ours} \\
\midrule
scan24 & 4.45 & 4.19 & 5.24 & 3.47 & 4.77 & \cellcolor{yellow}{2.99} & 4.07 & 5.99 & 3.16 & 4.20 & \cellcolor{orange}{2.24} & \cellcolor{red}{1.31} \\
scan37 & 4.67 & 3.85 & 5.09 & \cellcolor{yellow}{3.61} & 7.61 & 3.77 & 4.33 & 5.26 & 4.36 & 5.05 & \cellcolor{orange}{3.43} & \cellcolor{red}{3.14} \\
scan40 & 2.51 & \cellcolor{yellow}{2.26} & 3.99 & \cellcolor{orange}{2.10} & 5.34 & 3.01 & 2.54 & 4.13 & 2.59 & 3.47 & 2.77 & \cellcolor{red}{1.53} \\
scan55 & 1.90 & 2.49 & \cellcolor{yellow}{1.42} & \cellcolor{orange}{1.05} & 4.75 & 2.10 & 4.07 & 4.22 & 2.35 & 2.79 & 2.40 & \cellcolor{red}{0.92} \\
scan63 & 2.81 & 3.49 & 5.10 & \cellcolor{yellow}{2.37} & 4.07 & 2.98 & 3.05 & 3.14 & \cellcolor{orange}{1.88} & 3.05 & 3.71 & \cellcolor{red}{1.19} \\
scan65 & \cellcolor{yellow}{2.92} & 3.97 & 4.33 & \cellcolor{red}{1.38} & 4.15 & 4.10 & 3.61 & 5.00 & 3.28 & 3.16 & 3.22 & \cellcolor{orange}{1.98} \\
scan69 & 2.12 & \cellcolor{yellow}{1.91} & 5.36 & \cellcolor{orange}{1.41} & 7.38 & 2.73 & 3.16 & 4.20 & 2.32 & 3.16 & 2.49 & \cellcolor{red}{1.25} \\
scan83 & \cellcolor{yellow}{2.05} & 2.49 & 3.15 & \cellcolor{red}{1.85} & 6.00 & 2.38 & 3.29 & 4.78 & \cellcolor{orange}{1.88} & 3.17 & 3.34 & 2.16 \\
scan97 & 2.93 & 2.37 & 5.78 & \cellcolor{orange}{1.74} & 5.38 & 4.17 & 4.86 & 4.54 & \cellcolor{yellow}{1.87} & 3.62 & 2.71 & \cellcolor{red}{1.40} \\
scan105 & \cellcolor{yellow}{2.05} & 2.27 & 2.07 & \cellcolor{orange}{1.10} & 5.26 & 3.45 & 2.79 & 4.85 & 2.06 & 4.16 & 2.84 & \cellcolor{red}{1.07} \\
scan106 & \cellcolor{yellow}{2.01} & 2.90 & 2.79 & \cellcolor{orange}{1.46} & 5.36 & 4.40 & 5.24 & 5.44 & 2.25 & 3.09 & 2.69 & \cellcolor{red}{1.26} \\
scan110 & N/A & 4.60 & 5.73 & 2.28 & 5.51 & \cellcolor{yellow}{1.97} & 5.29 & 4.69 & \cellcolor{orange}{1.92} & 5.97 & 3.08 & \cellcolor{red}{0.95} \\
scan114 & \cellcolor{orange}{1.10} & 1.38 & \cellcolor{yellow}{1.20} & 1.25 & 4.65 & 2.53 & 2.41 & 7.82 & 1.40 & 2.49 & 2.20 & \cellcolor{red}{0.82} \\
scan118 & 2.72 & 2.57 & 5.64 & \cellcolor{orange}{1.44} & 5.63 & 2.94 & 3.01 & 4.36 & \cellcolor{yellow}{2.43} & 3.23 & 2.52 & \cellcolor{red}{1.18} \\
scan122 & \cellcolor{yellow}{1.64} & 1.76 & 6.20 & \cellcolor{orange}{1.45} & 5.99 & 2.88 & 3.57 & 6.78 & 1.94 & 1.90 & 2.51 & \cellcolor{red}{1.07} \\
\midrule
Average & 2.56 & 2.84 & 4.21 & \cellcolor{orange}{1.86} & 5.46 & 3.09 & 3.69 & 5.01 & \cellcolor{yellow}{2.38} & 3.50 & 2.81 & \cellcolor{red}{1.42} \\
\midrule
\small{Training Data} & - & - & - & - & 74K & 1.5M & 0.6M & 8M & 27.1K & 35.8K & 8.5M & 86.4K \\
\bottomrule
\end{tabular}
\vspace{-9pt}
\label{tab:dtu-all}
\end{table*}
\begin{figure}
    \centering
    \includegraphics[width=\linewidth]{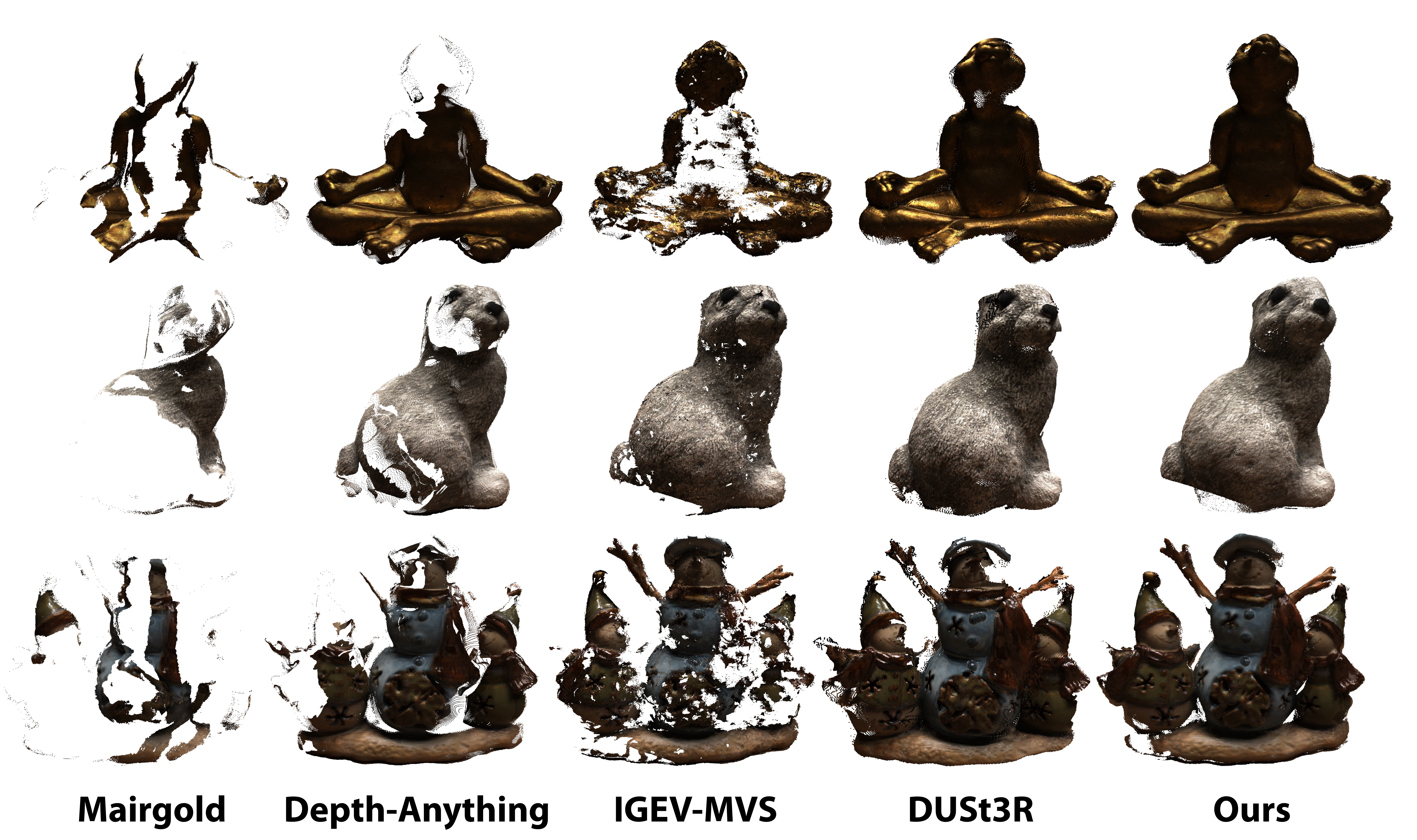}
    \caption{\textbf{Qualitative comparison on DTU dataset.} Except for DUSt3R which directly infers the point cloud, all other methods use Fusibile \cite{galliani2015massively} to fuse depth maps, filtering out points with inconsistencies across different viewpoints. \methodname produces a more complete final point cloud due to its relatively accurate multi-view consistent depth estimation. Please refer to the supplementary materials for results on other datasets.}
    \vspace{-5pt}
    \label{fig:dtu_comp}
\end{figure}

\subsubsection{Metrics}
Given our focus on multi-view geometric reconstruction, we primarily evaluate reconstruction quality.
We report Chamfer distance on DTU following \cite{yu2022monosdf}, F-score with a threshold of 5cm on ScanNet and Replica following \cite{yu2022monosdf}.
As for Waymo, we report the RMSE (root mean square error) between predicted depth maps and the provided lidar depth in the image space to represent geometry quality due to the lack of 3D Ground Truth geometry, following \cite{guo2023streetsurf}.

\subsection{Comparison}

\methodname is compared with recent methods on monocular depth estimation, depth completion, and multi-view reconstruction, which are introduced in the supplementary material.
We report both qualitative and quantitative comparisons against baselines, showcasing the reconstructed geometry in \Cref{fig:dtu_comp} and evaluating the reconstruction results on DTU in \Cref{tab:dtu-all}.
Please refer to more quantitative results on street and aerial scenes in the supplementary material, as well as additional qualitative results (including predicted depth maps and reconstructed geometry).
It is evident that \methodname consistently outperforms the baseline approaches across various types of real-world datasets.
While Marigold and Depth-Anything produce visually appealing depth visualizations, the depth values are often inaccurate on many datasets, leading to poor reconstruction results.
Metric3D, trained on a vast amount of real-world streetscape data, performs well on the Waymo dataset but shows poor performance on other datasets. Despite aligning with SfM depth on these datasets, the results remain low quality.
MVSNet and IGEV-MVS, although trained on DTU, exhibit only moderate performance when tested on DTU with inputs of three views.
NeuralRecon and SimpleRecon, both trained on ScanNet, demonstrate good results when evaluated on ScanNet, but we find that they nearly fail to work on other datasets.

\subsection{Ablation studies}

\subsubsection{Trade-off between precision and speed}

\methodname incorporates several factors that impact speed and reconstruction quality: 1) the number of denoising steps, 2) the number of ensemble size, 3) the implementation of additional alignment. To analyze the influence of these factors, we conduct a series of experiments on Replica dataset, the results are shown in \Cref{tab:ablation-speed}.
For fairness, all tests are conducted on a single A6000 GPU with the batch size as 1. The actual speed of \methodname can be further improved by increasing the batch size.
For the experiment where $N_{step}=1$, we first distill our model using latent consistency model (LCM) (details in supplementary materials) to enable it to achieve satisfactory results with just one step.

The results indicate that: 1) with the LCM-distilled model, effective depth estimation can be achieved with just one denoising step, significantly reducing denoising time without sacrificing much reconstruction quality.; 2) reducing the ensemble size also considerably enhances speed with minimal compromise in reconstruction quality; 3) The time consumed by alignment is minimal, while we also demonstrate that since \methodname can predict depth maps at the same scale as the conditioned sparse depth maps, the alignment step is not indispensable. In summary, \methodname offers a flexible trade-off between speed and reconstruction quality, allowing for adjustments based on specific application requirements.

\begin{table}[t]
\centering
\setlength{\tabcolsep}{0.5pt}
\captionof{table}{\textbf{Trade-off between precision and speed.} We evaluate the inference speed (in seconds) of each view (at the resolution of 680x1200) and final reconstruction quality of \methodname under different settings. This allows us to select the most suitable setting based on specific application requirements.}
\begin{tabular}{cccccccc}
\toprule
$N_{Step}$ & $N_{Ensemble}$ & Alignment & Inference time $\downarrow$ & F-Score $\uparrow$ \\
\midrule
10 & 5 & RANSAC       & 12.166 & 0.853 \\
10 & 1 & RANSAC       & 2.969  & 0.850 \\
1  & 5 & RANSAC       & 3.202  & 0.829 \\
1  & 1 & RANSAC       & 0.840  & 0.828 \\
1  & 1 & Least Square & 0.836  & 0.812 \\
1  & 1 & No Alignment & 0.829  & 0.780 \\
\bottomrule
\end{tabular}
\vspace{-12pt}
\label{tab:ablation-speed}
\end{table}

\subsubsection{SfM method}

\begin{figure}
    \centering
    \includegraphics[width=1.\linewidth]{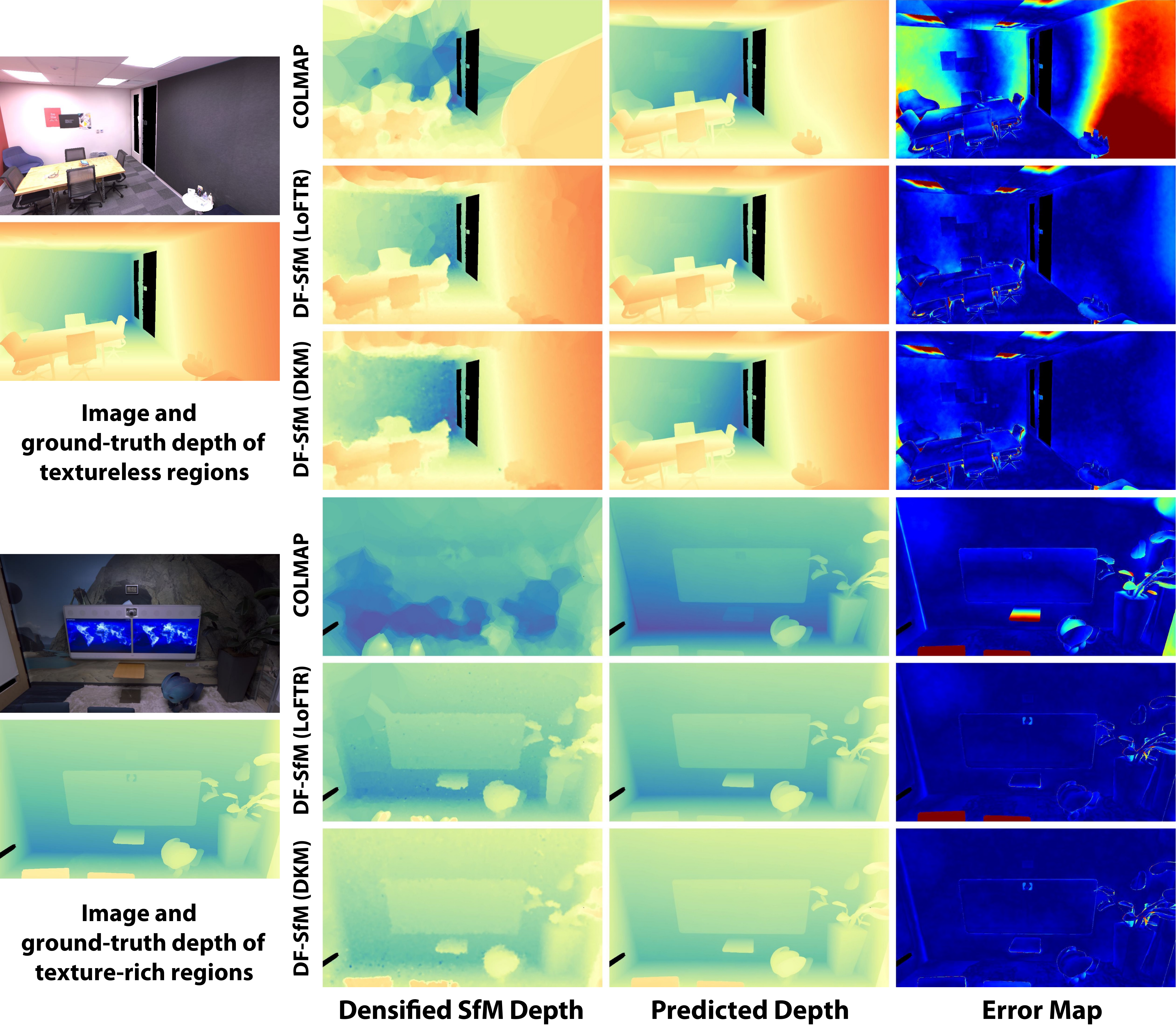}
    \caption{\textbf{Qualitative ablations of SfM methods.} 
    We conduct ablations on both texture-rich and textureless scenes.
    }
    \label{fig:ablation-sfm}
    \vspace{-20pt}
\end{figure}

Our training data are all processed by DF-SfM~\cite{he2024dfsfm} for sparse reconstruction, primarily due to DF-SfM's robustness in scenes with low textures.
During inference, other SfM methods can be utilized.
Note that our experiments on Urbanscene3D are based on COLMAP rather than DF-SfM.
To further analyze the impact of the choice of SfM method on our performance, we conduct an ablation study on Replica, which is an indoor dataset and contain large areas of low texture in some scenes, the results are shown in \Cref{tab:ablation-sfm} and \Cref{fig:ablation-sfm}.
During the construction of training data, DF-SfM uses LoFTR as its matcher, and for inference, good results can be achieved whether we continue using LoFTR or switch to DKM.
Meanwhile, the performance of \methodname based on COLMAP depends on specific characteristics of the scene. In regions with weak textures, COLMAP's keypoints tend to be extremely sparse and noisy, as a result, while the predicted depth map may look visually appealing, it often contains substantial errors. However, in regions with relatively rich textures, although COLMAP's keypoints are sparser than DF-SfM, they remain within an acceptable range and have good accuracy, resulting in high-quality depth maps generated by \methodnamenoblank.
In overall, \methodname is robust to the choice of SfM method: as long as the SfM can produce a reasonably good sparse point cloud for the scene, we can achieve high-quality depth estimation and reconstruction results based on it.

\vspace{-5pt}

\begin{table}[t]
\centering
\setlength{\tabcolsep}{3pt}
\captionof{table}{\textbf{Quantitative ablations of SfM methods.}  While \methodname is trained with the sparse points from DF-SfM (LoFTR), it is capable of directly utilizing the condition from COLMAP and DF-SfM (DKM) for inference, demonstrating the robustness.}
\begin{tabular}{lcccccc}
\toprule
& COLMAP & DF-SfM (LoFTR) & DF-SfM (DKM) \\
\midrule
F-score & 0.645 & \textbf{0.853} & 0.842 \\
\bottomrule
\end{tabular}
\vspace{-6pt}
\label{tab:ablation-sfm}
\end{table}

\subsubsection{Depth conditioning}
\label{sec:depth-conditioning}

To introduce SfM priors into depth estimation, we first densify the SfM-projected depth using KNN, then pass it through an encoder and concatenate it with a downsampled distance map as condition.
To analyze the impact of this design, we conduct ablation studies on KNN and the distance map on Replica dataset.
As shown in the \Cref{tab:ablation-knn}, the performance is poor without KNN, suggesting that original sparse depth map is not suitable for directly inputting into the encoder. A slight improvement is observed with $k=3$ compared to $k=1$, indicating that $k=3$ provides more information. Furthermore, whether at $k=1$ or $k=3$, omitting the distance map as a condition significantly worsens the results, demonstrating that the distance map effectively conveys crucial information about which pixels are originally valued and which are interpolated.

\begin{table}[t]
\centering
\setlength{\tabcolsep}{3pt}
\captionof{table}{\textbf{Ablation studies of depth conditioning.} We evaluate the impact on performance when using different values of $k$ in KNN (where $k=0$ indicates no densification) and the effect of using the distance map as condition.}
\begin{tabular}{lcccccc}
\toprule
$k$            & 0 & 0 & 1 & 1 & 3 & 3 \\
distance map & \ding{56} & \ding{52} & \ding{56} & \ding{52} & \ding{56} & \ding{52} \\
\midrule
F-score & 0.543 & 0.528 & 0.783 & 0.840 & 0.753 & \textbf{0.853} \\
\bottomrule
\end{tabular}
\vspace{-12pt}
\label{tab:ablation-knn}
\end{table}

\section{Conclusion}
We propose a new paradigm for multi-view geometric reconstruction of 3D scenes based on monocular depth estimation. 
Traditional routines for image-based 3D reconstruction rely on either multi-view stereo matching or differentiable inverse rendering which may suffer from high computational cost and low reconstruction quality caused by insufficient textures or very sparse input views.
Recent advances in large vision models have shown remarkable capabilities to recover high-quality depth maps from single images, but the estimated depth values are often imprecise and inconsistent across different views due to the inherent single-view ambiguity.  
We propose to incorporate SfM point clouds as a condition into the diffusion-based depth estimation process, thus naturally enhancing the metric accuracy and multi-view consistency of the estimated depth maps, which are then directly fused to reconstruct the scene geometry. 
We evaluate the reconstruction quality in various types of real-world datasets including indoor scenes, streetscapes, and aerial views, surpassing state-of-the-art MVS and implicit neural reconstruction-based methods.

\methodname also has some known limitations.
First, we require running SfM, which is infeasible in extreme cases where only two views with minimal overlap are available, making it nearly impossible to estimate camera poses or obtain a sparse point cloud, thereby preventing further reconstruction.
Additionally, \methodname is designed to reconstruct static scenes and is hard to handle moving elements.
Future work could explore integrating DUSt3R-like approaches to address more extreme cases and incorporating tracking methods to handle dynamic scenes.

\paragraph{Acknowledgement.}
This work was partially supported by NSFC (No. U24B20154), Zhejiang Provincial Natural Science Foundation of China (No. LR25F020003), and Information Technology Center and State Key Lab of CAD\&CG, Zhejiang University.

{
    \small
    \bibliographystyle{ieeenat_fullname}
    \bibliography{main}
}

\newpage
\appendix
\section*{Appendix}
\begin{figure*}[t]
    \centering
    \includegraphics[width=\linewidth]{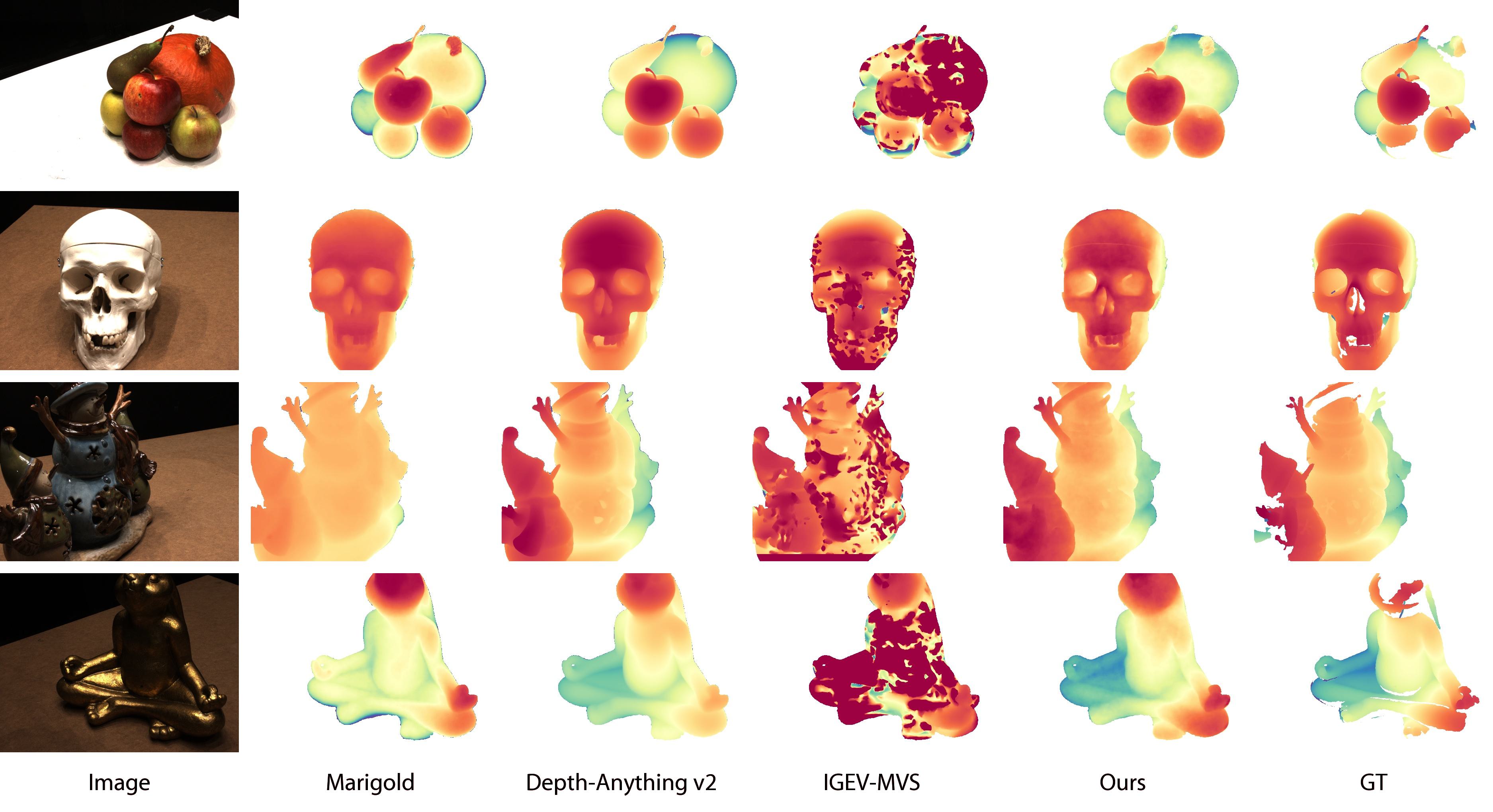}
    \caption{Qualitative comparison of depth estimation on DTU~\cite{jensen2014large}.}
    \label{fig:dtu_depth}
\end{figure*}

\begin{figure*}[t]
    \centering
    \includegraphics[width=\linewidth]{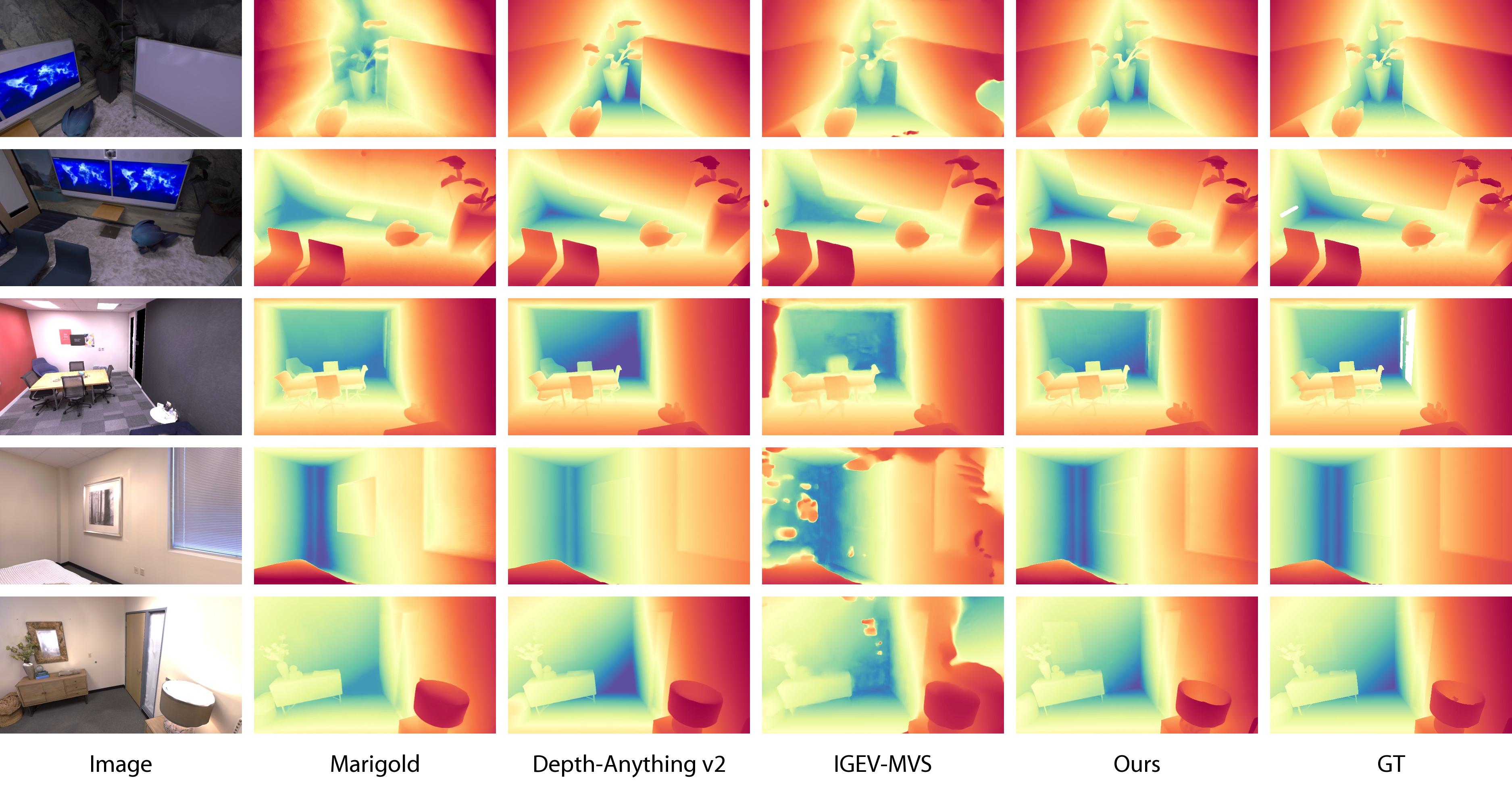}
    \caption{Qualitative comparison of depth estimation on Replica~\cite{replica19arxiv}.}
    \label{fig:replica_depth}
\end{figure*}

\begin{figure*}[t]
    \centering
    \includegraphics[width=\linewidth]{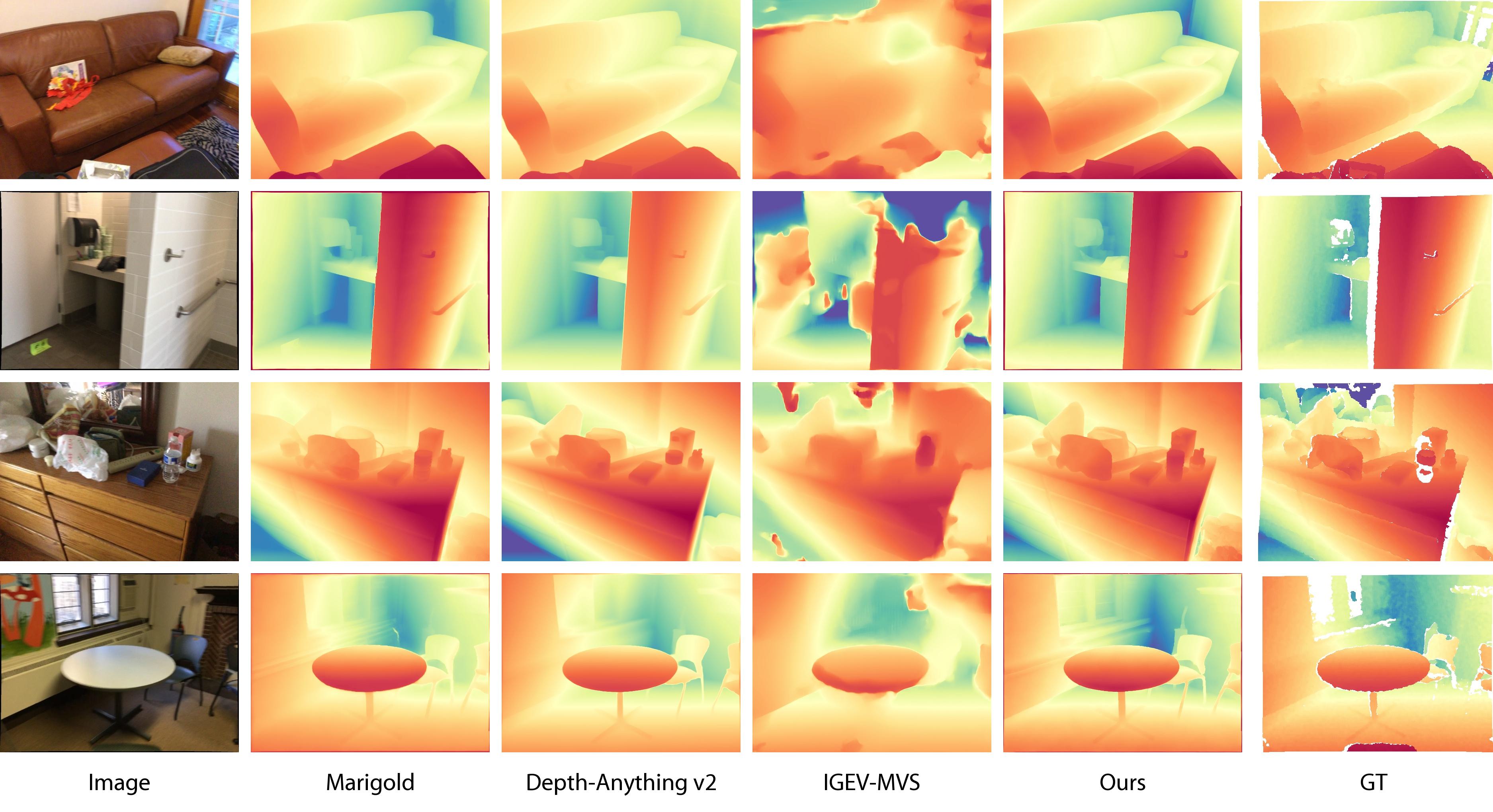}
    \caption{Qualitative comparison of depth estimation on ScanNet~\cite{dai2017scannet}.}
    \label{fig:scannet_depth}
\end{figure*}

\begin{figure*}[t]
    \centering
    \includegraphics[width=\linewidth]{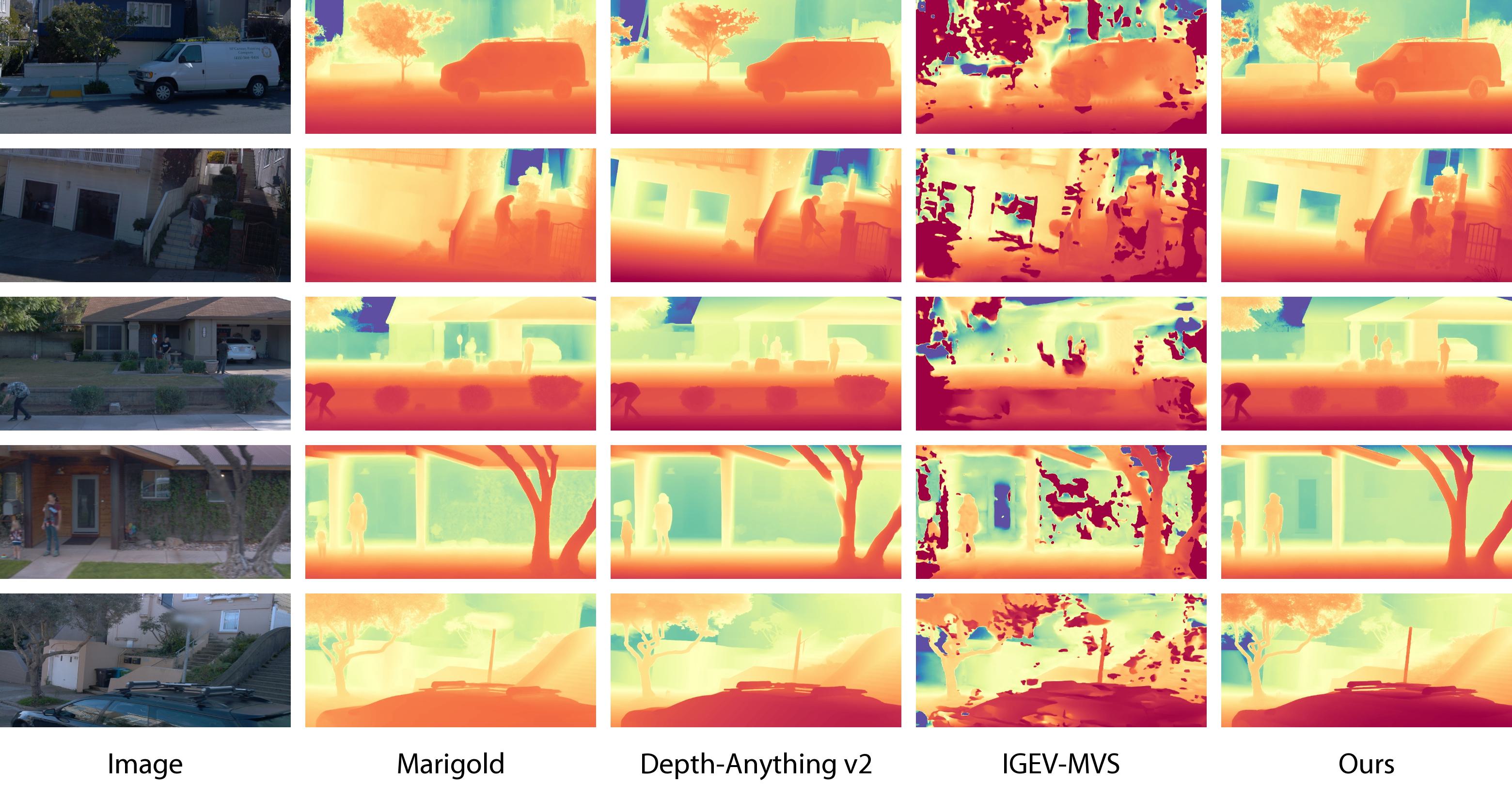}
    \caption{Qualitative comparison of depth estimation on Waymo~\cite{Sun_2020_CVPR}.}
    \label{fig:waymo_depth}
\end{figure*}

\begin{figure*}[t]
    \centering
    \includegraphics[width=\linewidth]{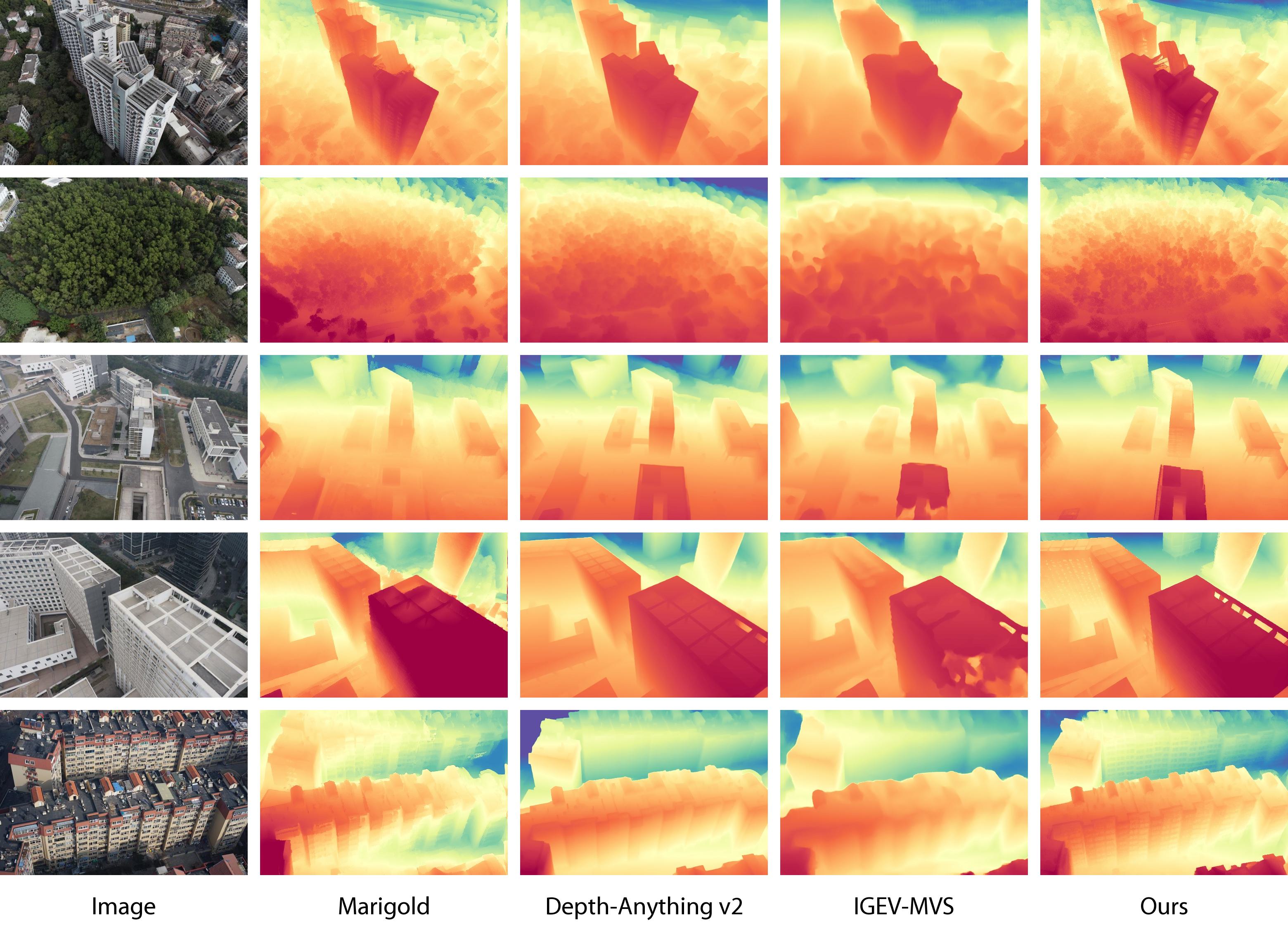}
    \caption{Qualitative comparison of depth estimation on UrbanScene3D~\cite{UrbanScene3D}.}
    \label{fig:log_depth}
\end{figure*}

\begin{figure*}[t]
    \centering
    \includegraphics[width=\linewidth]{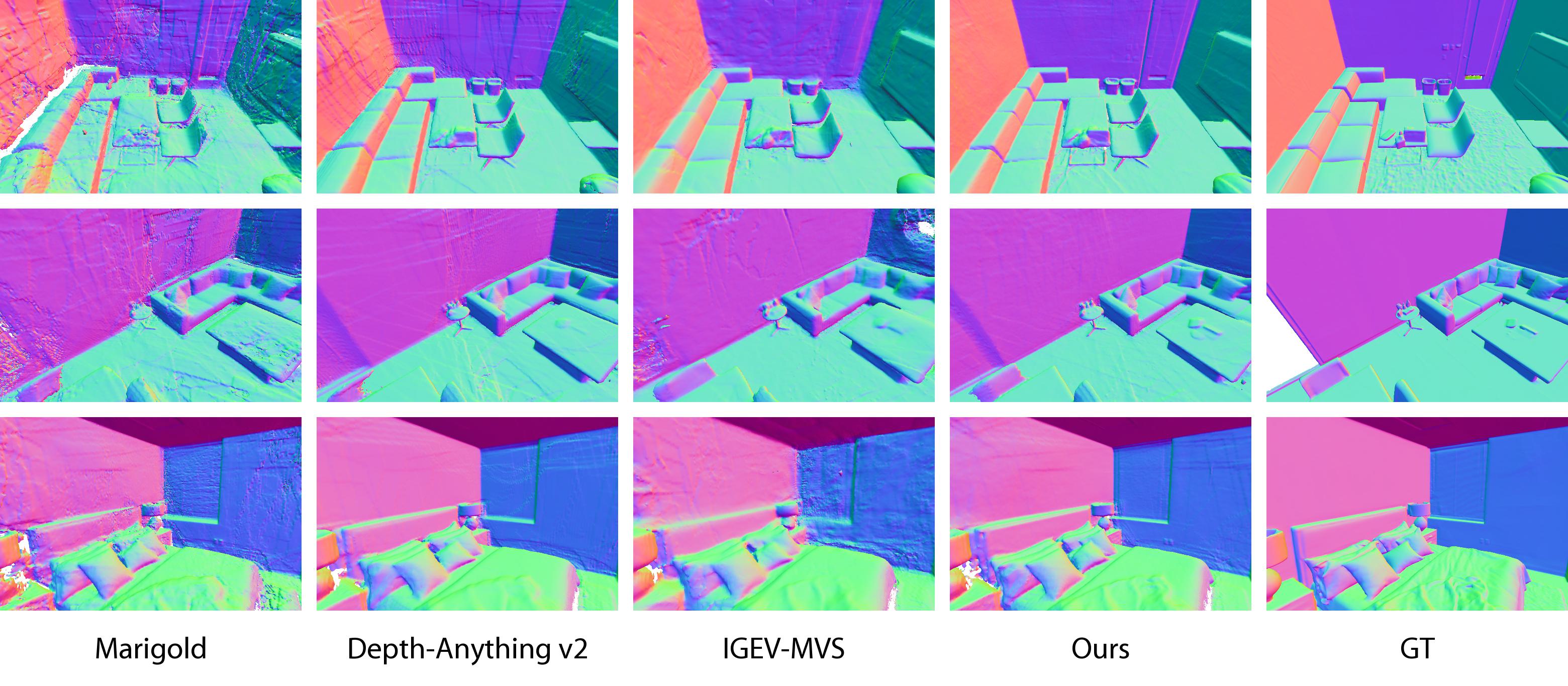}
    \caption{Qualitative comparison of geometric reconstruction on Replica~\cite{replica19arxiv}.}
    \label{fig:replica_mesh}
\end{figure*}

\begin{figure*}[t]
    \centering
    \includegraphics[width=0.82\linewidth]{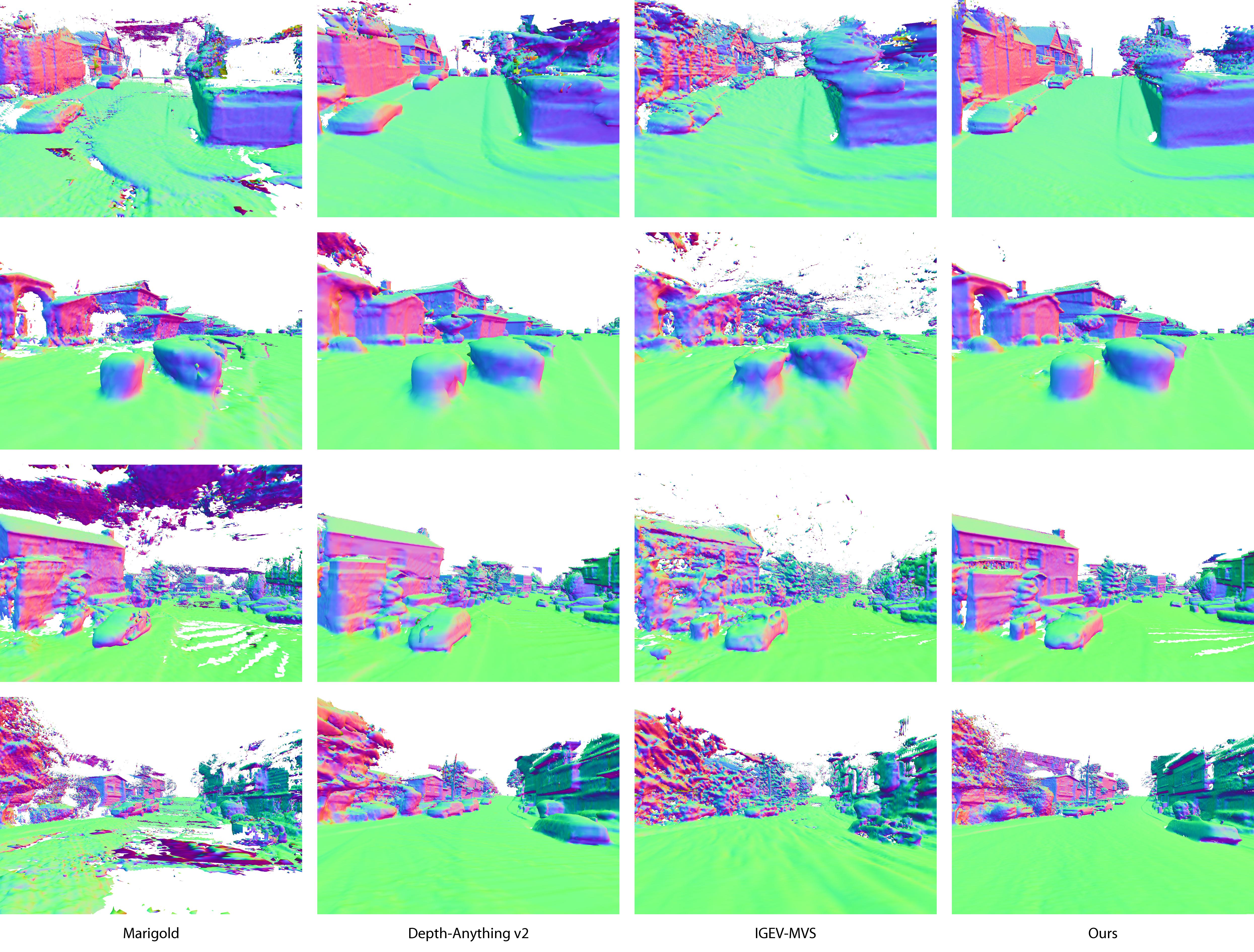}
    \caption{Qualitative comparison of geometric reconstruction on Waymo~\cite{Sun_2020_CVPR}.}
    \label{fig:waymo_mesh}
\end{figure*}

\begin{figure*}[t]
    \centering
    \includegraphics[width=0.82\linewidth]{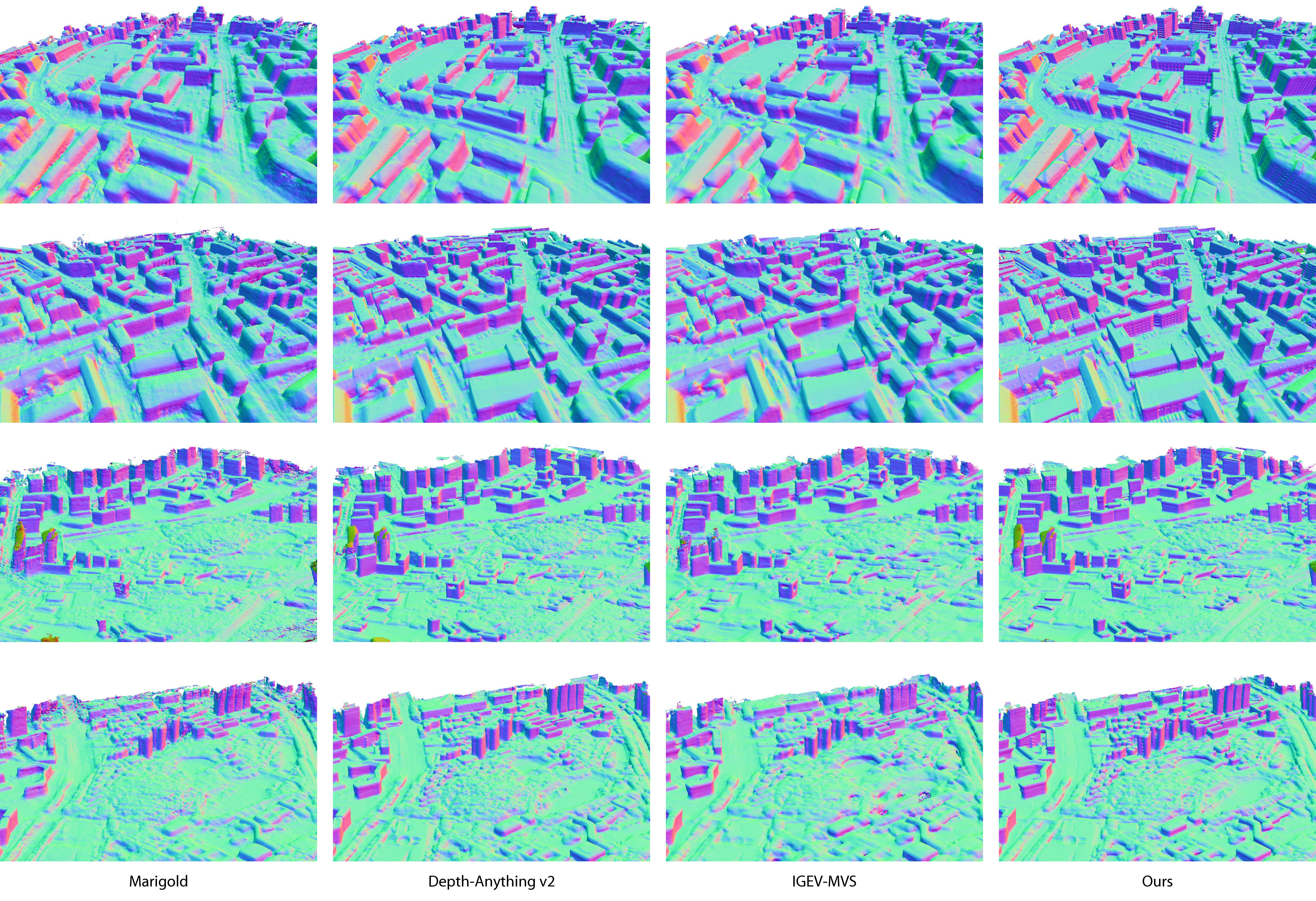}
    \caption{Qualitative comparison of geometric reconstruction on UrbanScene3D~\cite{UrbanScene3D}.}
    \label{fig:log_mesh}
\end{figure*}

\begin{figure*}[h]
    \centering
    \includegraphics[width=0.7\linewidth]{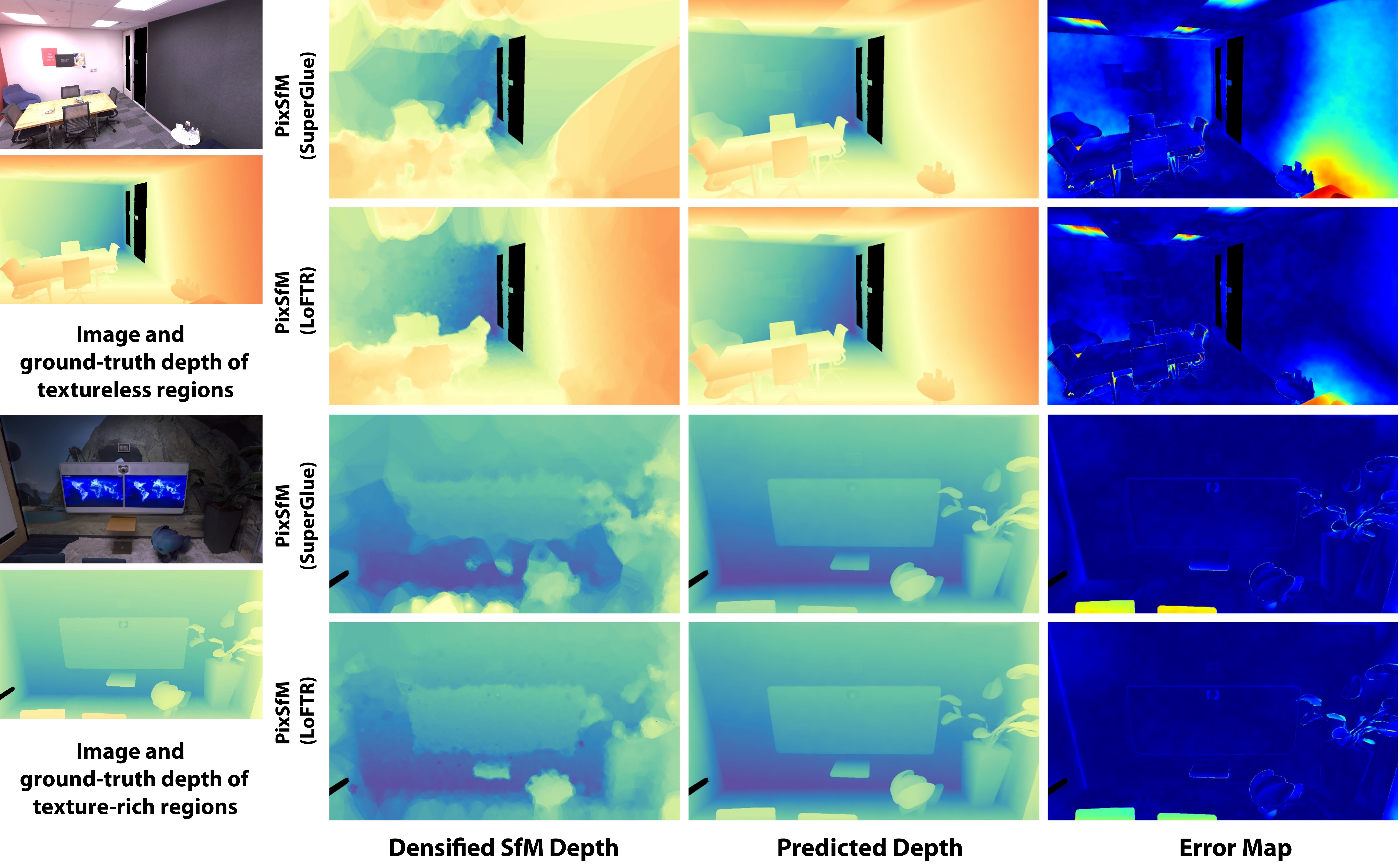}
    \caption{Results of our method based on PixSfM.}
    \label{fig:ablation-pixsfm}
\end{figure*}

\begin{figure*}[h]
    \centering
    \includegraphics[width=\linewidth]{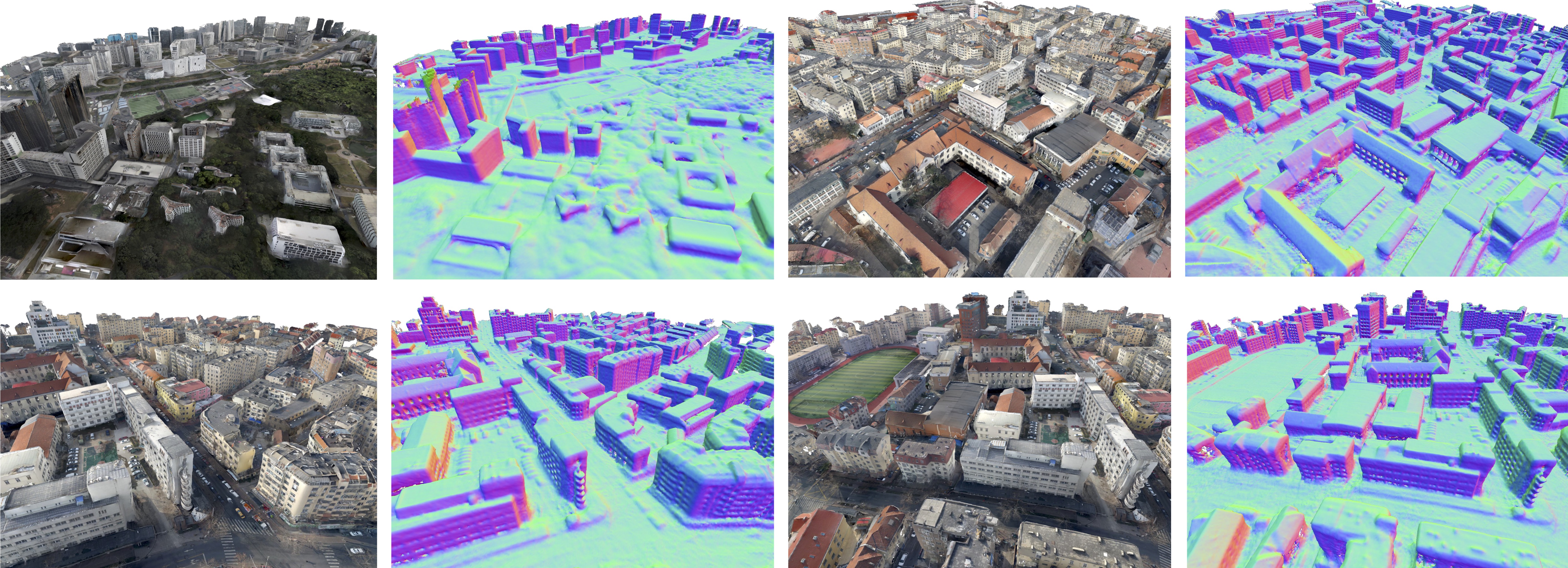}
    \caption{Visualization of our results with texture from texture mapping on UrbanScene3D~\cite{UrbanScene3D}.}
    \label{fig:results-urban}
\end{figure*}

{
\linespread{0.5}

\section{Baselines}
We compare our method with the following baseline methods in four categories:
\begin{itemize}
    \item \textit{Monocular depth estimation}: Marigold~\cite{ke2023repurposing}, Depth-Anything~\cite{yang2024depth} and Depth-Anything v2~\cite{yang2024depthv2} are monocular relative depth estimation methods. Due to discrepancies
    between their results and the true scale, we align their predictions with SfM depth before evaluation. Metric3D~\cite{yin2023metric3d} is a monocular metric depth estimation method.
    \item \textit{Depth completion:} SparseDC~\cite{long2024sparsedc} is currently the state-of-the-art method for monocular depth completion. We input RGB images and SfM depth maps from each frame of the test scenes into SparseDC for comparison.
    \item \textit{Optimization-based reconstruction:} MonoSDF~\cite{yu2022monosdf} and StreetSurf~\cite{guo2023streetsurf} model scenes using Signed Distance Fields (SDF), optimize the SDF through differentiable rendering, and leverage monocular geometric cues to enhance reconstruction quality.
    \item \textit{Learning-based MVS:} MVSNet~\cite{yao2018mvsnet}, IGEV-MVS~\cite{xu2023iterative} and SimpleRecon~\cite{sayed2022simplerecon} construct a cost volume from multi-view inputs to predict depth. NeuralRecon~\cite{sun2021neuralrecon} aggregates multi-view features in world coordinates to predict TSDF volumes, thereby extracting scene geometry. Dust3R~\cite{wang2023dust3r} uses a ViT model to reconstruct point maps from input image pairs.
\end{itemize}
For each depth estimation based method, we employ the same multi-view fusion technique as ours.

\section{Comparison results}
In addition to the geometric reconstruction results on DTU presented in the main paper, we provide more qualitative and quantitative comparisons and analyses in the supplementary material. These include qualitative comparisons of depth maps (in \Cref{fig:dtu_depth,fig:replica_depth,fig:scannet_depth,fig:waymo_depth,fig:log_depth}), qualitative comparisons of geometric reconstructions (in \Cref{fig:replica_mesh,fig:waymo_mesh,fig:log_mesh}), and quantitative evaluations (in \Cref{tab:waymo-all,tab:replica-all,tab:scannet-all}). When visualizing depth maps, we normalize all methods using the same range and employ the Spectral colormap for consistent visualization. Based on these results, we draw the following conclusions:
\begin{itemize}
    \item \textit{Monocular relative depth estimation methods:} These methods, particularly Depth-Anything v2, exhibit visually impressive depth predictions. However, their numerical accuracy is not as strong, as evidenced by several observations. First, their quantitative evaluation results are not particularly high. Second, their reconstructed meshes exhibit some noise, often caused by inconsistencies between different views. Lastly, the color differences between their depth map visualizations and the ground truth in some areas also reflect numerical errors.
    \item \textit{Monocular metric depth estimation methods:} Metric3D performs well on datasets like ScanNet and Waymo, partly because its training data includes real-world indoor and street-view data that closely resemble these scenes. However, Metric3D performs poorly on object-level and aerial datasets like DTU and UrbanScene3D.
    \item \textit{Depth completion:} SparseDC is not robust to real SfM depth inputs, which often contain noise, resulting in suboptimal depth completion and final reconstruction results.
    \item \textit{Optimization-based reconstruction methods:} These methods (e.g., MonoSDF, StreetSurf) achieve high-quality reconstructions in indoor scenes but suffer from very slow optimization processes. Moreover, their performance is less competitive in large-scale street-view scenes due to limited expressiveness.
    \item \textit{Learning-based MVS methods:} IGEV predicts relatively accurate depth maps in areas with enough views and rich textures. However, on DTU, its performance is hindered by the limited number of views, leading to suboptimal matching. While IGEV performs well overall in indoor and outdoor scenes, it struggles in low-texture and boundary regions. NeuralRecon and SimpleRecon achieve good results on ScanNet, however, we found that they perform very poorly on other datasets.
    \item \textit{Our method:} \methodname not only produces visually pleasing depth predictions but also achieves higher numerical accuracy. It is robust in low-texture regions and performs well across various datasets, demonstrating consistent and reliable results.
\end{itemize}

\section{Implementation of LCM}

We analyse the trade-off between speed and reconstruction quality in the main paper, where we distill our model using Latent Consistency Model (LCM)~\cite{luo2023latent} to reduce the number of denoising steps. Specifically, we fix the UNet in the original model as the teacher UNet and use it to initialize the student UNet and the target UNet. During training, the student UNet is optimized using consistency objective, while the target UNet updates its parameters via exponential moving average (EMA). During inference, the trained LCM enables few-step denoising, achieving satisfactory results even with a single step.

\section{Additional ablation study on sfm method}
Without any retraining, we directly evaluate our performance using PixSfM with two different matchers: SuperPoint+SuperGlue and LoFTR, as shown in \Cref{fig:ablation-pixsfm}.

\section{Visualization with texture}
To better visualize the reconstruction results of our method, we apply an off-the-shelf texture mapping method~\cite{waechter2014let} to our meshes on the UrbanScene3D dataset. The results are presented in \Cref{fig:results-urban}.
}

\begin{table*}[t]
    \centering
    \setlength{\tabcolsep}{4pt}
    \captionof{table}{\textbf{Quantitative results on Waymo~\cite{Sun_2020_CVPR}.} The metrics for COLMAP, F2NeRF, and StreetSurf are sourced from the StreetSurf paper. Note that their evaluations are conducted in LiDAR space, whereas ours and other baselines are in image space. While the assessment results from both approaches should be closely aligned, they may not be identical. We report their metrics for reference.}
    \begin{tabular}{cccccccccccc}
    \toprule
    Sequence & \footnotesize{COLMAP} & \footnotesize{F2-NeRF} & \footnotesize{StreetSurf} & \footnotesize{Marigold} & \tiny{Depth-Anything} & \tiny{Depth-Anything v2} & \footnotesize{MVSNet} & \footnotesize{IGEV-MVS} & \footnotesize{Metric3D} & \footnotesize{SparseDC} & \footnotesize{Ours} \\
    \midrule
    seg1006130.. & 7.10 & 8.87 & 2.99 & 2.92 & \cellcolor{yellow}{2.42} & 2.76 & 15.53 & 4.13 & \cellcolor{orange}{2.29} & 7.93 & \cellcolor{red}{2.11} \\
    seg1027514.. & 7.47 & 16.52 & 2.91 & 2.95 & \cellcolor{yellow}{2.70} & 2.77 & 13.98 & 6.06 & \cellcolor{orange}{2.67} & 10.54 & \cellcolor{red}{2.47} \\
    seg1067626.. & 9.06 & 35.59 & \cellcolor{red}{4.34} & 5.74 & 5.29 & 5.50 & 13.95 & 9.55 & \cellcolor{orange}{4.57} & 17.36 & \cellcolor{yellow}{5.00} \\
    seg1137922.. & 12.39 & 20.10 & \cellcolor{yellow}{5.70} & 7.41 & 6.08 & 6.57 & 13.35 & 10.77 & \cellcolor{red}{5.01} & 16.85 & \cellcolor{orange}{5.61} \\
    seg1172406.. & 13.62 & 9.00 & 2.57 & 2.31 & \cellcolor{yellow}{1.88} & 2.16 & 16.73 & 4.15 & \cellcolor{red}{1.49} & 7.40 & \cellcolor{orange}{1.57} \\
    seg1287964.. & 10.34 & 6.73 & \cellcolor{orange}{3.19} & 3.59 & \cellcolor{yellow}{3.27} & 3.34 & 10.14 & 5.99 & 3.39 & 10.52 & \cellcolor{red}{3.05} \\
    seg1308545.. & 8.64 & 15.50 & 4.12 & 3.81 & \cellcolor{yellow}{3.52} & 3.66 & 14.15 & 6.12 & \cellcolor{red}{2.91} & 9.94 & \cellcolor{orange}{3.18} \\
    seg1314219.. & 6.75 & 19.30 & \cellcolor{orange}{3.48} & 4.27 & 3.82 & 3.81 & 12.61 & 7.18 & \cellcolor{yellow}{3.60} & 12.19 & \cellcolor{red}{3.28} \\
    seg1319679.. & 7.63 & 23.50 & 4.76 & 4.73 & \cellcolor{yellow}{4.31} & 4.45 & 14.46 & 5.58 & \cellcolor{red}{3.67} & 11.11 & \cellcolor{orange}{3.99} \\
    seg1323841.. & 7.32 & 20.19 & \cellcolor{orange}{3.13} & 3.57 & 3.47 & 3.44 & 12.88 & 6.74 & \cellcolor{yellow}{3.33} & 12.20 & \cellcolor{red}{2.95} \\
    seg1347637.. & 5.93 & 21.72 & \cellcolor{red}{1.84} & 2.74 & 2.75 & 2.66 & 17.72 & 2.54 & \cellcolor{yellow}{2.09} & 5.19 & \cellcolor{orange}{1.85} \\
    seg1400454.. & 8.08 & 39.85 & 3.29 & 2.89 & \cellcolor{yellow}{2.63} & 2.72 & 11.66 & 6.07 & \cellcolor{orange}{2.58} & 11.18 & \cellcolor{red}{2.43} \\
    seg1434813.. & 8.48 & 35.96 & \cellcolor{yellow}{4.74} & 5.93 & 6.19 & 6.20 & 17.82 & 6.05 & \cellcolor{orange}{4.50} & 10.79 & \cellcolor{red}{4.30} \\
    seg1442480.. & 7.85 & 36.35 & \cellcolor{orange}{2.97} & 3.70 & 3.40 & 3.40 & 12.92 & 7.06 & \cellcolor{yellow}{2.98} & 12.80 & \cellcolor{red}{2.96} \\
    seg1486973.. & 5.52 & 3.53 & 2.82 & 2.25 & \cellcolor{yellow}{1.72} & 2.10 & 18.86 & 3.15 & \cellcolor{orange}{1.70} & 6.71 & \cellcolor{red}{1.48} \\
    seg1506235.. & 7.84 & 27.61 & 2.40 & 2.36 & \cellcolor{yellow}{2.19} & \cellcolor{orange}{2.12} & 13.32 & 6.02 & 2.22 & 11.00 & \cellcolor{red}{1.83} \\
    seg1522170.. & 11.28 & 16.66 & \cellcolor{yellow}{4.87} & 5.49 & 5.30 & 5.58 & 17.53 & 6.75 & \cellcolor{orange}{4.61} & 11.72 & \cellcolor{red}{4.35} \\
    seg1527063.. & 2.62 & 7.82 & 1.98 & 1.80 & \cellcolor{yellow}{1.56} & 1.81 & 20.80 & 3.83 & \cellcolor{orange}{1.38} & 7.38 & \cellcolor{red}{1.32} \\
    seg1534950.. & 4.31 & 7.80 & \cellcolor{yellow}{2.56} & 2.94 & 2.67 & 2.80 & 14.00 & 4.48 & \cellcolor{orange}{2.29} & 7.53 & \cellcolor{red}{1.92} \\
    seg1536582.. & 6.57 & 10.41 & 2.47 & 1.94 & \cellcolor{yellow}{1.54} & 1.76 & 20.86 & 3.15 & \cellcolor{orange}{1.48} & 7.44 & \cellcolor{red}{1.46} \\
    seg1586862.. & 5.94 & 18.78 & \cellcolor{yellow}{2.60} & 3.16 & 2.98 & 3.14 & 14.71 & 5.45 & \cellcolor{orange}{2.53} & 8.64 & \cellcolor{red}{2.47} \\
    seg1634531.. & 5.31 & 11.85 & 2.23 & 2.33 & \cellcolor{yellow}{1.97} & 2.16 & 15.59 & 3.59 & \cellcolor{red}{1.53} & 7.54 & \cellcolor{orange}{1.79} \\
    seg1647019.. & 10.36 & 12.25 & 4.31 & 4.64 & \cellcolor{yellow}{4.20} & 4.27 & 14.20 & 7.28 & \cellcolor{orange}{3.88} & 12.05 & \cellcolor{red}{3.74} \\
    seg1660852.. & 5.11 & 4.72 & 3.91 & 3.50 & \cellcolor{yellow}{2.92} & 2.93 & 17.28 & 4.23 & \cellcolor{red}{2.62} & 7.95 & \cellcolor{orange}{2.68} \\
    seg1664636.. & 6.54 & 13.86 & \cellcolor{yellow}{2.26} & 2.53 & 2.54 & 2.61 & 17.42 & 4.04 & \cellcolor{orange}{1.94} & 7.45 & \cellcolor{red}{1.66} \\
    seg1776195.. & 14.52 & 25.24 & 3.90 & 4.22 & \cellcolor{yellow}{3.72} & 3.76 & 12.04 & 7.24 & \cellcolor{orange}{3.58} & 12.24 & \cellcolor{red}{3.56} \\
    seg3224923.. & 5.42 & 7.16 & 3.53 & 3.00 & \cellcolor{yellow}{2.43} & 2.72 & 14.79 & 4.49 & \cellcolor{red}{2.07} & 8.57 & \cellcolor{orange}{2.21} \\
    seg3425716.. & 18.81 & 30.68 & \cellcolor{orange}{3.00} & 3.67 & 3.20 & \cellcolor{yellow}{3.03} & 18.46 & 7.55 & 3.23 & 9.94 & \cellcolor{red}{2.95} \\
    seg3988957.. & 6.07 & 5.66 & 3.30 & 3.36 & \cellcolor{orange}{2.95} & \cellcolor{yellow}{2.98} & 12.66 & 5.78 & 3.07 & 10.91 & \cellcolor{red}{2.90} \\
    seg4058410.. & 5.46 & 7.02 & \cellcolor{yellow}{2.62} & 3.05 & 3.00 & 2.92 & 12.62 & 4.62 & \cellcolor{red}{2.37} & 8.24 & \cellcolor{orange}{2.48} \\
    seg8811210.. & 7.16 & 27.30 & 3.83 & 3.28 & \cellcolor{yellow}{2.94} & 3.04 & 16.42 & 6.40 & \cellcolor{orange}{2.75} & 10.75 & \cellcolor{red}{2.70} \\
    seg9385013.. & 9.10 & 49.34 & 4.52 & 5.03 & \cellcolor{yellow}{4.34} & 4.42 & 17.68 & 9.89 & \cellcolor{red}{4.08} & 14.63 & \cellcolor{orange}{4.33} \\
    \midrule
    Average & 8.08 & 18.65 & 3.35 & 3.60 & \cellcolor{yellow}{3.25} & 3.36 & 15.22 & 5.81 & \cellcolor{orange}{2.89} & 10.21 & \cellcolor{red}{2.83} \\
    \bottomrule
    \end{tabular}
    \label{tab:waymo-all}
    \end{table*}

\begin{table*}[h]
\centering
\setlength{\tabcolsep}{2.2pt}
\captionof{table}{Quantitative results on ScanNet~\cite{dai2017scannet}.}
\begin{tabular}{cccccccccccccc}
\toprule
& \scriptsize{COLMAP} & \tiny{Manhattan-SDF} & \scriptsize{MonoSDF} & \scriptsize{Marigold} & \tiny{Depth-Anything} & \tiny{Depth-Anything v2} & \scriptsize{Metric3D} & \scriptsize{SparseDC} & \scriptsize{NeuralRecon} & \scriptsize{SimpleRecon} & \scriptsize{MVSNet} & \scriptsize{IGEV-MVS} & \footnotesize{Ours} \\
\midrule
0050\_00 & 0.563 & 0.673 & \cellcolor{red}{-} & 0.669 & 0.674 & 0.669 & 0.507 & 0.250 & 0.661 & \cellcolor{yellow}{0.718} & 0.075 & 0.391 & \cellcolor{orange}{0.750} \\
0084\_00 & 0.631 & 0.630 & \cellcolor{red}{-} & 0.733 & 0.692 & \cellcolor{yellow}{0.863} & 0.516 & 0.204 & 0.805 & \cellcolor{orange}{0.881} & 0.066 & 0.600 & 0.732 \\
0580\_00 & 0.590 & 0.632 & \cellcolor{red}{-} & 0.627 & \cellcolor{yellow}{0.688} & 0.644 & 0.441 & 0.245 & 0.484 & 0.542 & 0.137 & 0.512 & \cellcolor{orange}{0.720} \\
0616\_00 & 0.365 & 0.472 & \cellcolor{red}{-} & 0.543 & \cellcolor{orange}{0.597} & 0.578 & 0.437 & 0.179 & 0.518 & 0.590 & 0.077 & 0.430 & \cellcolor{yellow}{0.596} \\
\midrule
Average & 0.537 & 0.602 & \cellcolor{red}{0.733} & 0.643 & 0.663 & \cellcolor{yellow}{0.689} & 0.475 & 0.220 & 0.617 & 0.683 & 0.089 & 0.483 & \cellcolor{orange}{0.700} \\
\bottomrule
\end{tabular}
\label{tab:scannet-all}
\end{table*}
\begin{table*}[t]
\centering
\setlength{\tabcolsep}{5pt}
\captionof{table}{Quantitative results on Replica~\cite{replica19arxiv}.}
\begin{tabular}{cccccccccc}
\toprule
 & \footnotesize{MonoSDF} & \footnotesize{Marigold} & \tiny{Depth-Anything} & \tiny{Depth-Anything v2} & \footnotesize{Metric3D} & \footnotesize{SparseDC} & \footnotesize{MVSNet} & \footnotesize{IGEV-MVS} & \footnotesize{Ours} \\
\midrule
room\_1 & \cellcolor{red}{-} & 0.61 & 0.79 & \cellcolor{orange}{0.86} & 0.77 & 0.25 & 0.54 & \cellcolor{yellow}{0.84} & 0.84 \\
office\_0 & \cellcolor{red}{-} & 0.52 & 0.69 & 0.71 & 0.47 & 0.29 & 0.73 & \cellcolor{yellow}{0.85} & \cellcolor{orange}{0.90} \\
office\_2 & \cellcolor{red}{-} & 0.58 & 0.52 & 0.62 & 0.57 & 0.23 & 0.56 & \cellcolor{yellow}{0.78} & \cellcolor{orange}{0.82} \\
\midrule
Average & \cellcolor{red}{0.86} & 0.57 & 0.67 & 0.73 & 0.61 & 0.25 & 0.61 & \cellcolor{yellow}{0.82} & \cellcolor{orange}{0.85} \\
\bottomrule
\end{tabular}
\label{tab:replica-all}
\end{table*}

\end{document}